\documentclass{article}

\usepackage{arxiv}

\usepackage[utf8]{inputenc} 
\usepackage[T1]{fontenc}    
\usepackage{hyperref}       
\usepackage{url}            
\usepackage{booktabs}       
\usepackage{amsfonts}       
\usepackage{nicefrac}       
\usepackage{microtype}      
\usepackage{graphicx}
\usepackage[numbers,square,comma]{natbib}
\usepackage{doi}

\usepackage{amsmath,amssymb}               
\usepackage{multirow}                      
\usepackage{enumerate}                     
\usepackage[labelformat=empty]{subfig}     
\usepackage{algorithm}                     
\usepackage{algpseudocode}                 

\usepackage{anyfontsize}
\usepackage{colortbl}   
\usepackage{xcolor}     

\usepackage{steinmetz}

\usepackage{tikz}
\usetikzlibrary{mindmap}
\usetikzlibrary{calc,positioning}
\usepackage{lipsum,adjustbox}
\usetikzlibrary{decorations.pathmorphing}
\usetikzlibrary{arrows.meta,positioning,fit}

\definecolor{lightblue}{HTML}{E8F4F8}
\definecolor{darkblue}{HTML}{003366}
\definecolor{lightgreen}{HTML}{E8F5E9}
\definecolor{darkgreen}{HTML}{2E7D32}
\definecolor{lightyellow}{HTML}{FFF9C4}

\newlength{\colfigwidth}
\usepackage{siunitx} 

\usepackage{xurl}

\title{A ROS2 based Human-In-The-Loop Framework for UAV Search and Rescue Operations}

\date{}

\newcommand{\orcidicon}[1]{\href{#1}{\includegraphics[height=7bp]{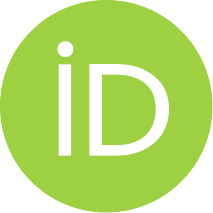}}}
\newcommand{\authsep}{\ \ $|$\ \ }
\newcommand{\authorblock}{%
\parbox{\dimexpr\textwidth-2\tabcolsep\relax}{\centering
Anubhav Pandey\textsuperscript{1}\hskip1bp\orcidicon{https://orcid.org/0009-0001-3517-2975}\authsep
Harshit Nagpal\textsuperscript{1}\hskip1bp\orcidicon{https://orcid.org/0009-0000-8089-1701}\authsep
Aditya Agarwal\textsuperscript{1}\hskip1bp\orcidicon{https://orcid.org/0009-0005-9084-1093}\authsep
Tejasvi Alladi\textsuperscript{1}\hskip1bp\orcidicon{https://orcid.org/0000-0003-4612-3180}\authsep
Fei Richard Yu\textsuperscript{2}\hskip1bp\orcidicon{https://orcid.org/0000-0003-1006-7594}\par
\vspace{6pt}
{\normalfont\small
\textsuperscript{1}Department of Computer Science and Information Systems, Birla Institute of Technology and Science, Pilani, Pilani Campus, Vidya Vihar, Pilani, Rajasthan 333031, India\par
\textsuperscript{2}Department of Systems and Computer Engineering, Carleton University, Ottawa, K1S 5B6, Ontario, Canada\par
\vspace{6pt}
\textbf{Correspondence:}~Tejasvi Alladi (\href{mailto:tejasvi.alladi@pilani.bits-pilani.ac.in}{tejasvi.alladi@pilani.bits-pilani.ac.in})\authsep Anubhav Pandey (\href{mailto:p20220020@pilani.bits-pilani.ac.in}{p20220020@pilani.bits-pilani.ac.in})\par}%
}}
\author{\authorblock}

\renewcommand{\shorttitle}{A ROS2 based Human-In-The-Loop Framework for UAV Search and Rescue Operations}

\hypersetup{
pdftitle={A ROS2 based Human-In-The-Loop Framework for UAV Search and Rescue Operations},
pdfauthor={Anubhav Pandey, Harshit Nagpal, Aditya Agarwal, Tejasvi Alladi, Fei Richard Yu},
pdfkeywords={Human-Robot Interaction, ROS2, UAV Path Guidance, Gesture Recognition, Depth Estimation, Search and Rescue, Gazebo Simulation},
}

\begin{document}
\maketitle

\begin{abstract}
Existing Unmanned Aerial Vehicle (UAV) autonomy frameworks lack human contextual awareness, limiting real-time adaptability in unstructured Search and Rescue (SAR) environments. To address data availability, we propose a Robot Operating System (ROS2)-based human-in-the-loop (HITL) framework that translates human directional gestures (left/right) into UAV navigation commands using RGB-Depth (RGB-D) sensing and YOLOv11 (You Only Look Once). The deprecation of ROS1 necessitates this migration to ROS2, prioritising real-time communication, scalability, and long-term maintainability. Unlike approaches optimised for peak accuracy under ideal conditions, the proposed system prioritises robustness and real-time deployment in degraded SAR environments. Our system achieves 89\% confidence in depth estimation under occlusion and varying environmental conditions in Gazebo-PX4 simulations, a critical factor for UAVs that enables situational awareness during SAR operations. In foggy conditions, RGB-D fusion consistently provides more informative cues than RGB alone, with an average improvement of 46.7\% across all scenarios and distances. The framework further demonstrates superior occlusion handling (+15\%), adaptive RGB-D fusion, temporal consistency modelling, and real-time deployment via a ROS2 data distribution service (DDS), with end-to-end latency below 50 ms. We further implement the modified YOLOv11 and the depth-weighted pose estimation and evaluate them on a held-out simulated test set and on real people from the public UAV-GESTURE dataset, in clear air and in Koschmieder fog. Without retraining, the framework reads the gestures of real people with 80\% accuracy, depth keeps 77\% at a visibility of 10~m where RGB drops to 24\%, and the modified YOLOv11 reaches 91\% in simulation and 99\% on held-out real people when trained on real data. The open-source ROS2 implementation provides a structured pathway for SAR-UAV platforms transitioning from ROS1, ensuring seamless integration of human intent. This work bridges critical gaps in human-robot collaboration, enabling UAVs to act directly on operator gestures without complex path planning, thus advancing mission-critical ROS2 deployments in SAR operations.
\end{abstract}


\section{Introduction}\label{sec:intro}
Unmanned Aerial Vehicles (UAVs) have become a game changer in search-and-rescue (SAR) operations \cite{de_alcantara_andrade_autonomous_2019}. Their ability to quickly provide aerial surveillance in disaster zones, collapsed buildings, and remote terrain is unparalleled. However, while autonomous UAVs perform exceptionally well in controlled environments, their reliance on pre-programmed flight paths and sensor-based navigation can limit their flexibility in dynamic SAR scenarios \cite{liu_real-time_2021}. In these real-world situations, conditions are constantly changing -- shifting debris, victims may be hidden from view, and new obstacles can appear without warning. This is where human judgement becomes more valuable than rigid, pre-programmed algorithms. For example, a firefighter controlling a UAV inside a smoke-filled building may need to override the system quickly, but traditional joystick interfaces can struggle to provide the precision required in high-stress moments.

\subsection{The Need for Human-in-the-Loop (HITL) Systems}

HITL systems aim to combine human expertise with machine autonomy, but current solutions often come with challenges such as high latency and operator overload \cite{cacace_control_2016}. As user interaction plays a critical role in operational efficiency, voice commands, while intuitive, tend to be unreliable in noisy environments, and touchscreen interfaces pull the operator’s attention away from more critical responsibilities at the worst possible moment. Gesture-based control offers a promising alternative, allowing human operators to interact with UAVs through simple, natural hand gestures---such as \textit{halt} or \textit{move left}---which require minimal training \cite{yoo_motion_2022}. This provides a more seamless, intuitive connection between the operator and the UAV, enabling faster, more effective responses when it matters most.

\subsection{The ROS2 Imperative for SAR Robotics}

\begin{figure}[ht]
\centerline{\includegraphics[width=\colfigwidth]{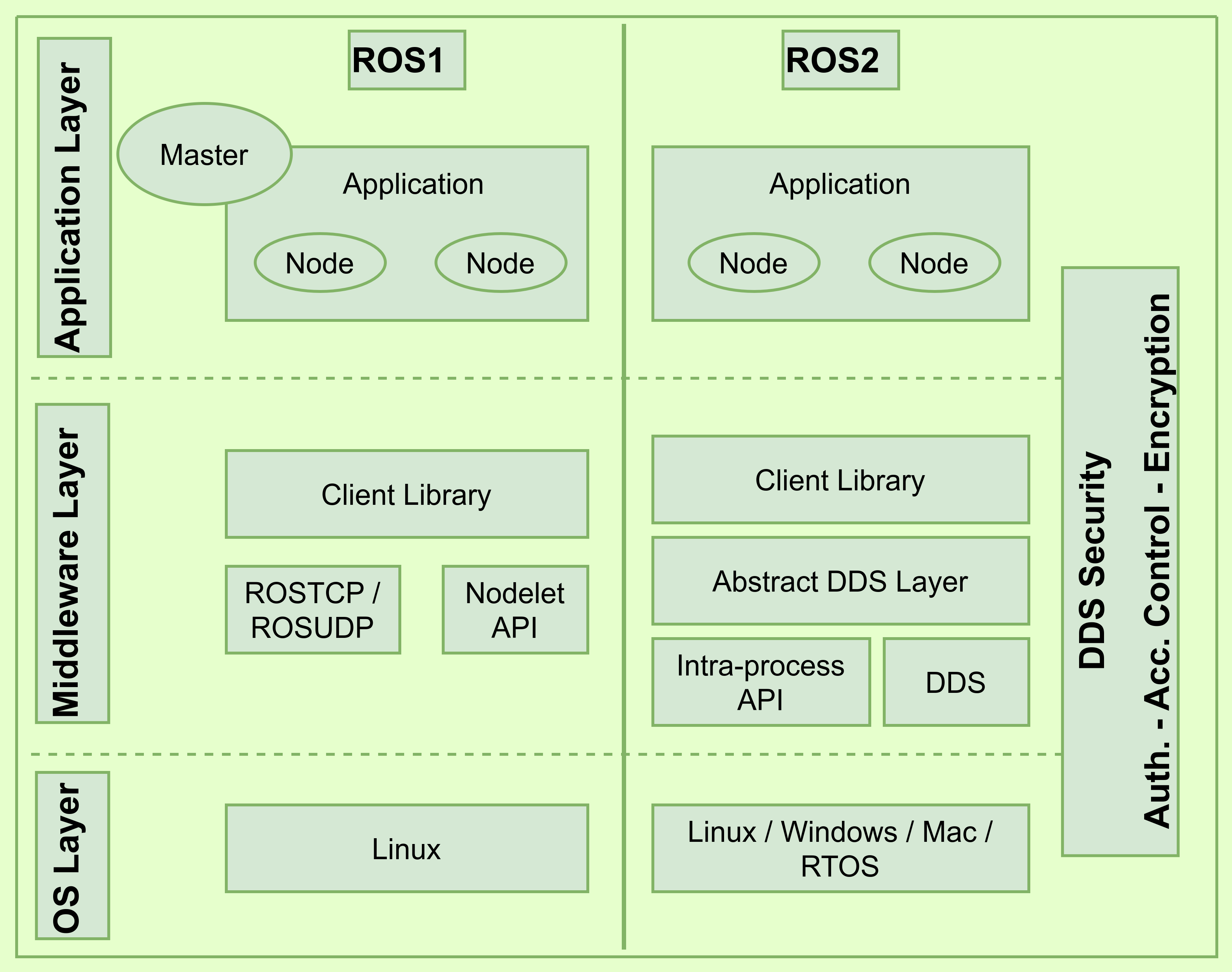}}
\caption{ROS1 vs ROS2. Redrawn and adapted from Maruyama et al.~\cite{maruyama_exploring_2016}.\label{fig:ros2}}
\end{figure}

Robot Operating System (ROS) \cite{quigley_ros_2009} is an open-source framework for building robotics applications more efficiently. It is a middleware, a set of tools, libraries, and conventions that simplify communication between hardware and software components in a robot or a drone. In ROS, topics act as communication channels for nodes, allowing the broadcast of information that multiple other nodes can receive. A node is an executable that communicates with other nodes via ROS by publishing or subscribing to topics. ROS has two versions: ROS1 and ROS2 \cite{macenski_robot_2022}. While ROS1 laid the foundation for robotic communication with its publisher-subscriber model, its real-time performance, security, and scalability limitations make it less ideal for mission-critical SAR operations. Referring to Fig.~\ref{fig:ros2}, ROS2, however, addresses these challenges head-on with several important features \cite{ramezani_human-centric_2024, faessler_autonomous_2016,kratky_autonomous_2021,probine_shared_2021,jarecki_mixed_2024,djeradi_toward_2021} :

\begin{itemize}
    \item \textbf{Secure, Priority-Based Communication:} By integrating the Data Distribution Service (DDS), ROS2 ensures secure, low-latency transmission of gesture commands.
    \item \textbf{Deterministic Execution:} With thread-safe executors and lock-free memory management and real-time optimisations, ROS2 ensures high reliability, especially in real-time tasks like obstacle avoidance and path planning, even in complex and dynamic environments.
    \item \textbf{High-Fidelity Simulation Support:} ROS2's integration with Gazebo allows for thorough testing in simulated SAR environments, ensuring optimal UAV performance before deployment.
    \item \textbf{Improved Build System and Scalability:} Introduced a more modern and flexible build system and tooling ecosystem, such as \textit{``ament''} and \textit{``colcon''}.

    \item \textbf{Modular Architecture:} The flexible middleware and language-agnostic communication in ROS2 make it an ideal framework for seamlessly integrating components, such as a YOLOv11-based gesture recognition module (in Python).
    \item \textbf{Cross-Platform and Embedded Support:} The supported platforms for ROS2 are Ubuntu, Debian, Windows, Mac OS, etc. Also, Micro-ROS extends ROS2 to microcontrollers, enabling lightweight embedded applications.
\end{itemize}

These advancements make ROS2 the ideal framework \cite{bonci_robot_2023,pandey_autonomous_2025} for developing next-generation HITL systems for SAR UAVs, where real-time responsiveness and security are critical.

\subsection{Contributions}

This paper introduces a ROS2-based framework using Gazebo simulator, PX4 Autopilot and YOLOv11 algorithm that integrates gesture-driven control for SAR UAVs. The following are the main contributions of this paper:

\begin{itemize}
    \item \textbf{ROS2 Migration Support:} We propose a modular, hardware-agnostic algorithmic framework to simplify the transition from ROS1 to ROS2 for SAR-UAV developers. The framework outlines the integration of gesture recognition nodes, PX4-compatible flight control, providing a clear conceptual basis for implementation and field deployment.
    \item \textbf{Unified Vision-Based Gesture Pipeline for UAV Control:} We design a unified pipeline that combines RGB and depth data to estimate human pose, calculate distance, and classify pointing gestures using YOLOv11, enabling intuitive HITL UAV guidance in complex SAR missions.
    \item \textbf{Robustness in Cluttered and Degraded Environments:} We deploy a YOLOv11 model on RGB and depth data to detect directional gestures, which enhances recognition accuracy in cluttered environments. Furthermore, we quantify the framework's resilience in degraded visual conditions, demonstrating a 46.7\% average improvement in detection confidence under fog compared to RGB-only baselines, ensuring operational viability where conventional vision fails.

    \item \textbf{Mission Efficiency Validation:} Through Gazebo simulations with PX4 autopilot, we demonstrate that the framework's direct gesture-to-command mapping minimises computational overhead, which is critical for time-sensitive SAR operations.    
    \item \textbf{Extended Evaluation on Datasets:} We implement the modified YOLOv11 (YOLOv11-$\mathcal{L}_{total}$) and the depth-weighted pose estimation, and test them together with Algorithm~\ref{alg:algo} on a fixed simulated test set and on real people from the public UAV-GESTURE dataset, in clear air and in two fog models (Section~\ref{sec:ext_eval}). As the framework is modular, its perception model, gesture rules and UAV commands can be exchanged for other UAV tasks and evaluated on other datasets in the same way.
\end{itemize}

\section{Related Works}
\label{sec:lit_sec}

\subsection{Human-UAV Interaction}  

Human-UAV interaction \cite{lygouras_unsupervised_2019} has evolved from basic static gestures in controlled settings to more sophisticated multimodal systems designed for dynamic environments. Andrade et al. \cite{de_alcantara_andrade_autonomous_2019} planned the paths of cooperative SAR UAVs in real time without operator input, and Yoo et al. \cite{yoo_motion_2022} controlled a UAV in real time with hand gestures, combining a thumb-mounted IMU with a camera. While effective at reducing latency, such vision-based gesture approaches relied heavily on RGB cameras, making them vulnerable to occlusion and varying lighting conditions. Liu et al. \cite{liu_real-time_2021} extended this concept by integrating deep learning for gesture recognition in rescue missions. However, because they relied on 2D skeletal tracking, they were prone to perspective distortions, particularly in high-altitude UAV operations where perception accuracy is critical.  

Further advancements were made by Cacace et al. \cite{cacace_control_2016}, who introduced multimodal interaction for controlling multiple UAVs. Although this method expanded control capabilities, it lacked depth-sensing technology, limiting its effectiveness in cluttered SAR environments. To overcome these limitations, our framework combines depth cameras with RGB data, which significantly enhances robustness against occlusion, lighting variations, and limited situational awareness. By incorporating 3D spatial information, our system demonstrates strong performance in dynamic SAR environments, surpassing approaches that rely exclusively on RGB-based recognition \cite{de_alcantara_andrade_autonomous_2019, liu_real-time_2021, cacace_control_2016}.  

\subsection{ROS2 for Human-UAV Interaction}  

ROS1 has long been the dominant middleware for UAV applications due to its flexibility and community-driven adoption. However, its architecture lacks real-time determinism and struggles with mission-critical responsiveness. Sampedro et al. \cite{sampedro_fully-autonomous_2019} deployed a fully autonomous indoor SAR UAV and tested it in Gazebo with PX4 SITL and in real flights, but it executes its missions without human intervention. Similarly, McConkey et al. \cite{mcconkey_semi-autonomous_2023} introduced a low-cost semi-autonomous ArduPilot UAV for wilderness SAR operations, but its video is transferred manually after the flight for human detection with YOLOv3, so the operator cannot act on detections during the mission.  

Recent efforts to transition UAV control to ROS2 have focused on improving human interaction reliability. Ramezani et al. \cite{ramezani_human-centric_2024} emphasised trajectory planning optimised for human comfort but treated autonomy and human input as independent processes rather than a collaborative system. Our work builds on these advancements by leveraging ROS2's DDS middleware to ensure deterministic communication and secure human-UAV interaction. By prioritising gesture commands with sub-\SI{50}{\milli\second} latency, we keep the operator in the loop during the mission, which the systems above do not support \cite{sampedro_fully-autonomous_2019, mcconkey_semi-autonomous_2023, ramezani_human-centric_2024}. This approach lets the operator redirect the UAV with a gesture at any time, ensuring UAV efficiency in time-sensitive SAR missions. 

In recent years, significant research has centred on classification methods for human activity and gesture recognition in UAV-based search and rescue (SAR) operations, seeking to enable intuitive human–drone interaction without relying on complex path-planning strategies. Lee et al. \cite{lee_deep_2022} proposed a deep learning–based posture recognition and tracking system that leverages OpenPose for 2D skeleton extraction and a YOLOv3‐supported classifier to distinguish seven human postures—such as punch, kick, squat, and wave—in real time, achieving high robustness under environmental uncertainties (viewpoint, illumination, occlusion). Building on multi‐view fusion, Saini et al. \cite{saini_airpose_2022} introduced AirPose, which combines aerial imagery from multiple camera angles to estimate 3D human pose and shape, improving classification accuracy for complex movement patterns in SAR scenarios.

Chang Liu et al. \cite{liu_real-time_2021} curated a ten‐gesture ``body rescue'' dataset (e.g., Kick, Punch, Squat, Attention, Cancel) recorded from onboard UAV cameras; their system introduced novel dynamic Attention and Cancel gestures to control UAV proximity and reset commands, reporting 99.8\% testing accuracy in gesture recognition and 94.7\% on hand gestures. Comprehensive surveys have contextualised these advances. However, the limitations are the drone positioning and hovering in the difficult environmental conditions in the SAR scenario. Mliki et al.'s \cite{mliki_human_2020} study on UAV‐captured video sequences categorises state‐of‐the‐art algorithms for aerial human action recognition, highlighting the challenges of arbitrary viewpoints and rapid motion for accurate classification. Rahman et al. \cite{rahman_comprehensive_2024} reviewed machine learning approaches for UAV detection and classification, emphasising the need for robust feature extraction under varying flight altitudes and environmental conditions. More recently, a new Drone-Action dataset \cite{perera_drone-action_2019} introduced 13 dynamic human actions, including boxing, jogging and waving, in 240 outdoor video clips recorded from a low-flying drone, underscoring the need for aerial data with sufficient body detail for action recognition in real‐world SAR operations. Beyond deep learning, unsupervised detection frameworks have been explored: an embedded vision system on a fully autonomous UAV demonstrated real-time human detection without operator supervision \cite{lygouras_unsupervised_2019}, suitable for rapid deployment in crisis zones.
Overall, these classification approaches lay the groundwork for gesture-based UAV control in SAR missions. By prioritising the recognition of human intent over navigation itself, they enable responsive, operator-driven actions in critical, time-sensitive environments.

\subsection{Bridging Simulation and Real-World Deployment}  

Despite advancements in shared control systems, many studies have been validated only in controlled environments, overlooking real-world SAR complexities. Probine et al. \cite{probine_shared_2021} investigated shared control of robotic airships in indoor experiments but did not address outdoor constraints such as urban debris and adverse weather conditions. Jarecki and Lee \cite{jarecki_mixed_2024} explored mixed-reality teleoperation for industrial robotic manipulators, neglecting UAV-specific challenges. Field-tested systems, such as the ArduPilot UAV of McConkey et al. \cite{mcconkey_semi-autonomous_2023}, have improved UAV validation but often exclude human interaction from testing frameworks.  

To bridge this gap, we validate our framework using realistic Gazebo simulations that replicate real-world SAR scenarios \cite{bianchi_novel_2023}, including urban rubble, collapsed buildings, and dense forest canopies. By integrating PX4 for real-time UAV control, our system is thoroughly tested in demanding environmental conditions. This extensive evaluation shows that the proposed framework meets the demands of operational SAR missions, offering a scalable and robust solution for next-generation UAV systems.

\subsection{Recent Advances in Embedded Depth-YOLO Systems and SAR-Specific Vision}

Recent developments in depth-aware object detection have focused on optimising embedded efficiency. Lugaresi et al. \cite{lugaresi_mediapipe_nodate} introduced MediaPipe, a framework for building perception pipelines that manages CPU and GPU resources and synchronises video streams for low-latency processing across mobile and desktop devices. This shows that real-time perception pipelines can run on resource-constrained hardware, although MediaPipe does not address depth sensing.

Parallel to detection efficiency, SAR-specific research has advanced significantly. An ensemble approach using YOLOv9 and EfficientDet (Basireddy et al. \cite{incoft25})  achieved 94.18\% accuracy on Heridal and Aerial Rescue Object Detection datasets, demonstrating the viability of deep learning for aerial human detection in authentic SAR scenarios. While their work focuses on human detection, our gesture-control approach tackles the complementary challenge of converting the operator’s intent into actionable UAV navigation commands.

Architectural innovations in YOLO for specific tasks demonstrate that domain-focused customisation yields meaningful improvements. Improved YOLOv11pose (Lei Liu et al. \cite{liu_improved_2024})  showed 2.1–2.2\% accuracy gains through DECA-Block attention mechanisms. Using depth as input gives a 15\% improvement in occluded forest scenarios (Table~\ref{tab:env_performance}) (vs. their 2.1\% from attention mechanisms alone), suggesting that adding depth values contributes substantially to occlusion robustness beyond architectural attention alone. In real-time pose estimation, transformer-based approaches like VTTransPose (Rui Li et al. \cite{li_efficient_2024}) achieved 76.5 AP on COCO datasets with 40\% memory reduction, demonstrating that computational efficiency and accuracy can be achieved simultaneously. Concurrently, research in real-time robotics (Guan and Sun, SIGBED \cite{guan2024ros2})  has identified temporal consistency in multi-sensor fusion as a critical requirement for safe operation, highlighting the importance of deterministic communication in mission-critical systems.

In the domain of direct UAV control, Abdalla and Baidya \cite{abdalla_uav_2025}  established a benchmark for vision-based gesture navigation, reporting 96.14\% detection accuracy in clear weather conditions using YOLOv4 and MediaPipe on edge computing platforms. A comprehensive survey by Bernardi et al. \cite{bernardi_comprehensive_2026} of image fusion methods categorises hundreds of fusion approaches and identifies feature-level fusion as the dominant strategy in modern deep learning systems, due to its balance between robustness and representational capacity compared to raw-data fusion.

Unlike these related approaches, our system uniquely: (1) quantifies fog robustness (46.7\% improvement over RGB-only), critical for smoke-filled rescue scenarios not evaluated in prior work; (2) implements gesture-to-command control enabling direct operator guidance rather than pure detection; and (3) deploys via ROS2 middleware for real-time deterministic guarantees addressing recent identified temporal consistency challenges in robotics.

{
\subsection{Multimodal Perception, Datasets and Fog Modelling}\label{sec:rw_ext}

Computer vision is a core capability of UAVs. Chen et al. \cite{chen_yolo-based_2023} reviewed the development and applications of YOLO-based UAV technology, and Katkuri et al. \cite{katkuri_autonomous_2024} found the YOLO framework in more than 39.5\% of 173 studies on deep-learning vision for autonomous UAVs. Ramachandran and Sangaiah \cite{ramachandran_review_2021} identified altitude, camera angle, occlusion and motion blur as the main challenges of UAV surveillance, and Shamitha and Ilango \cite{shamitha_applications_2024} reviewed deep-learning object detection for real-time drone surveillance. However, these reviews address detection rather than operator intent, which our gesture interpretation on the lightweight YOLOv11n-pose provides.

Gesture-based UAV control has mostly been tested in clear conditions. Liu and Szir\'anyi \cite{liu_real-time_2021} named extreme weather as a remaining limitation, Yoo et al. \cite{yoo_motion_2022} used ROS, Gazebo and PX4 SITL as we do, and Albright et al. \cite{albright_evaluating_2025} trained YOLOv11 on rendered gestures only.

Zou et al. \cite{zou_cma_2021} showed that concatenating or adding RGB and depth features ignores their different reliability and proposed cross-modal attention, which Hu et al. \cite{hu_cross-modal_2024} extended to multi-level features. However, both address objects rather than human gestures. Our modified YOLOv11 fuses depth into three feature-pyramid levels with cross-scale attention (Section~\ref{sec:t4_impl}), and Eq.~(\ref{eq:multimodal})--(\ref{eq:weights}) weight the estimates of each body joint.

YOLOv11-pose is pre-trained on the 17 COCO keypoints \cite{lin_microsoft_2014} used by Algorithm~\ref{alg:algo}. Perera et al. \cite{perera_uav-gesture_2019} published UAV-GESTURE, 119 UAV videos of 13 command gestures by eight participants, annotated with body joints but without depth, which we estimate with Depth-Anything-V2 \cite{yang_depth_2024} to test our framework on real people (Section~\ref{sec:real_eval}).

Fog is commonly modelled with Koschmieder's law \cite{koschmieder_theorie_1924, mccartney_optics_1976, narasimhan_vision_2002}, $I = J\,t + A\,(1-t)$, with transmission $t = e^{-\beta r}$ over the path length $r$, airlight $A$ and visibility $V = 3.912/\beta$. Sakaridis et al. \cite{sakaridis_semantic_2018} applied it to real images with estimated depth, as we do for the real videos (Section~\ref{sec:fog_models}). However, Bijelic et al. \cite{bijelic_seeing_2020} showed in real fog that cameras and active sensors degrade differently, so simulated fog only approximates real fog.
}

\section{System Architecture}
\label{sec:sys_arch}

\begin{figure}
\centerline{\includegraphics[width=\colfigwidth]{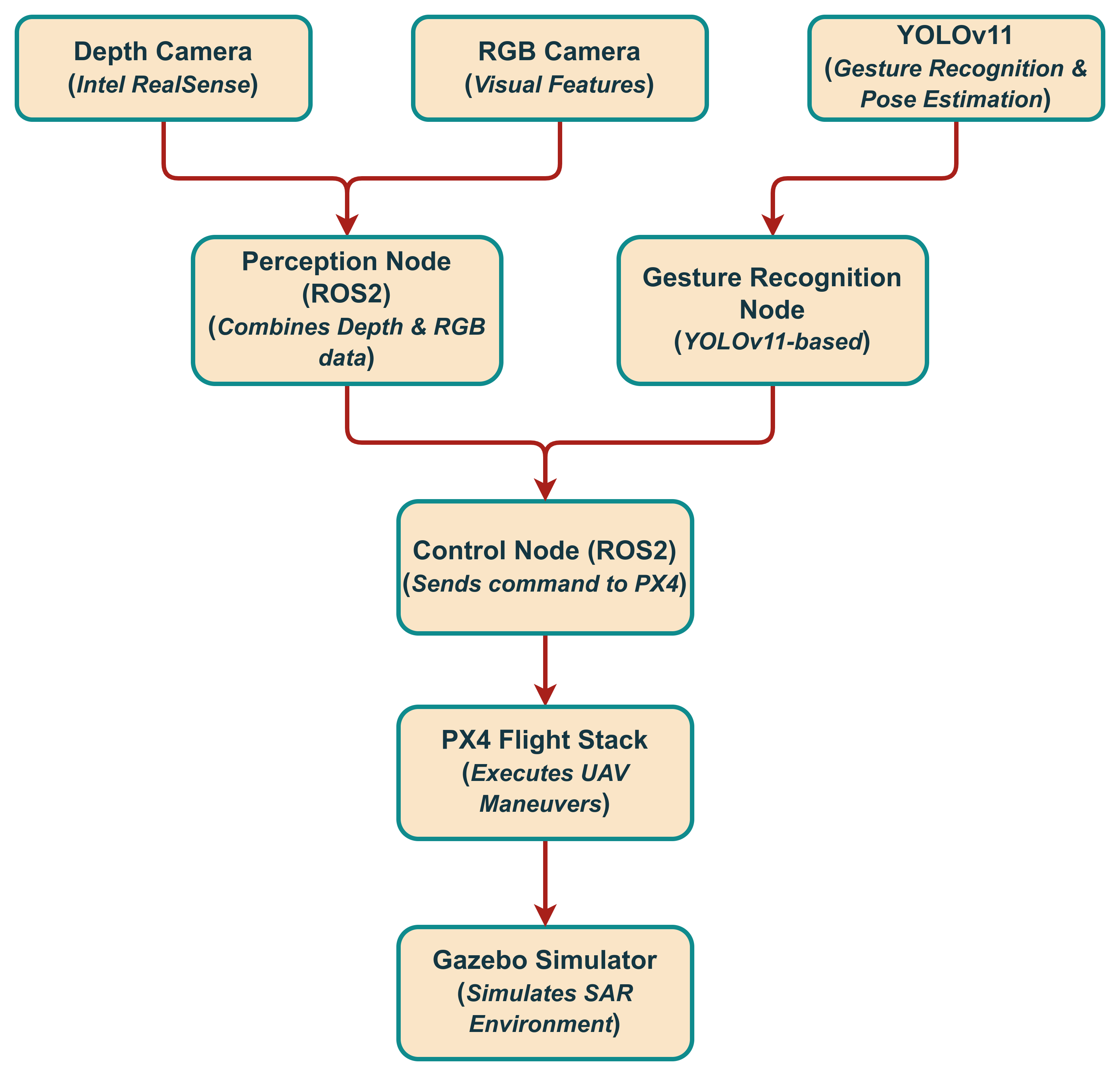}}
\caption{System Architecture \label{fig:system}}
\end{figure}

The proposed system architecture enables seamless, real-time human-UAV collaboration in search-and-rescue (SAR) scenarios through a tightly integrated sensing, perception, control, and simulation pipeline. Built on the Robot Operating System 2 (ROS2) framework, the architecture is designed for modularity, scalability, and real-time responsiveness. Fig.~\ref{fig:system} illustrates the overall system pipeline, where each module is described in detail below.

\subsection{Depth Camera (Intel RealSense)} The depth sensing module plays a critical role in extracting 3D spatial information from the environment. In our framework, the Intel RealSense depth camera collects real-time depth maps, providing detailed insight into the surrounding operational space. This data is essential for understanding obstacle proximity and estimating the human subject's location in three-dimensional space. The depth feed is especially crucial for accurately aligning the UAV's position relative to human gestures, enabling safe operation in cluttered or unstructured environments like forests or collapsed buildings \cite{noauthor_ros2_nodate}.
\subsection{RGB Camera (Visual Features)} Alongside the depth data, the RGB camera captures high-resolution colour images, which are crucial for recognising visual gestures and detecting human body keypoints. During perception, these visual cues are combined with depth data, ensuring that both spatial information and appearance-based features are available for further processing. The RGB feed provides context-aware information about the subject, including hand positions, pointing directions, and other semantic features that are not easily discernible from depth alone.
\subsection{YOLOv11 (Gesture Recognition \& Pose Estimation)} 
\begin{figure*}[tp]
\centering
\resizebox{\textwidth}{!}{%
\begin{tikzpicture}[
  font=\scriptsize, >=Latex, node distance=2mm,
  L/.style={draw=black!60, fill=black!7, rounded corners=1.5pt, align=left, minimum height=0.66cm, text width=3.45cm, inner sep=2.5pt},
  A/.style={draw=blue!70!black, fill=blue!8, rounded corners=1.5pt, align=left, minimum height=0.66cm, text width=3.45cm, inner sep=2.5pt},
  io/.style={draw=black!60, rounded corners=1.5pt, align=center, minimum height=0.66cm, inner sep=2.5pt},
  sum/.style={draw=blue!70!black, fill=blue!8, circle, inner sep=0pt, minimum size=0.36cm, font=\footnotesize},
  arr/.style={->, semithick, black!70},
  darr/.style={->, semithick, blue!70!black},
  sec/.style={font=\normalsize\bfseries}
]
\def\xb{5.1}
\node[io, text width=3.45cm] (in)  at (\xb, 0.9) {RGB channels $3\times640^2$};
\node[L] (b0)  at (\xb, 0)     {\textbf{0} Conv $3{\times}3$ s2 \hfill $16{\times}320^2$};
\node[L] (b1)  at (\xb,-0.9)   {\textbf{1} Conv $3{\times}3$ s2 \hfill $32{\times}160^2$};
\node[L] (b2)  at (\xb,-1.8)   {\textbf{2} C3k2 \hfill $64{\times}160^2$};
\node[L] (b3)  at (\xb,-2.7)   {\textbf{3} Conv $3{\times}3$ s2 \hfill $64{\times}80^2$};
\node[L] (b4)  at (\xb,-3.6)   {\textbf{4} C3k2 \hfill $128{\times}80^2$};
\node[sum] (s3) at (\xb,-4.4)  {$+$};
\node[L] (b5)  at (\xb,-5.2)   {\textbf{5} Conv $3{\times}3$ s2 \hfill $128{\times}40^2$};
\node[L] (b6)  at (\xb,-6.1)   {\textbf{6} C3k2 \hfill $128{\times}40^2$};
\node[sum] (s4) at (\xb,-6.9)  {$+$};
\node[L] (b7)  at (\xb,-7.7)   {\textbf{7} Conv $3{\times}3$ s2 \hfill $256{\times}20^2$};
\node[L] (b8)  at (\xb,-8.6)   {\textbf{8} C3k2 \hfill $256{\times}20^2$};
\node[L] (b9)  at (\xb,-9.5)   {\textbf{9} SPPF \hfill $256{\times}20^2$};
\node[L] (b10) at (\xb,-10.4)  {\textbf{10} C2PSA \hfill $256{\times}20^2$};
\node[sum] (s5) at (\xb,-11.2) {$+$};
\foreach \a/\b in {in/b0,b0/b1,b1/b2,b2/b3,b3/b4,b4/s3,s3/b5,b5/b6,b6/s4,s4/b7,b7/b8,b8/b9,b9/b10,b10/s5}
  \draw[arr] (\a) -- (\b);
\def\xd{0.6}
\node[io, text width=2.8cm] (din) at (\xd, 0.9) {Depth channel $D$\\ $1\times640^2$};
\node[A, text width=2.8cm] (st) at (\xd,-1.8)  {Stem: 3 Conv $3{\times}3$ \hfill $32{\times}160^2$};
\node[A, text width=2.8cm] (d3) at (\xd,-4.4)  {$\phi_3$: 2 Conv $3{\times}3$ \hfill $64{\times}80^2$};
\node[A, text width=2.8cm] (d4) at (\xd,-6.9)  {$\phi_4$: 2 Conv $3{\times}3$ \hfill $96{\times}40^2$};
\node[A, text width=2.8cm] (d5) at (\xd,-11.2) {$\phi_5$: 2 Conv $3{\times}3$ \hfill $128{\times}20^2$};
\draw[darr] (din) -- (st);
\draw[darr] (st) -- (d3);
\draw[darr] (d3) -- (d4);
\draw[darr] (d4) -- (d5);
\draw[darr] (d3) -- (s3) node[midway, above, align=center, font=\scriptsize] {$W_3$: $1{\times}1$, $64{\rightarrow}128$};
\draw[darr] (d4) -- (s4) node[midway, above, align=center, font=\scriptsize] {$W_4$: $1{\times}1$, $96{\rightarrow}128$};
\draw[darr] (d5) -- (s5) node[midway, above, align=center, font=\scriptsize] {$W_5$: $1{\times}1$, $128{\rightarrow}256$};
\def\xn{9.4}
\node[L] (n11) at (\xn,-11.2) {\textbf{11} Upsample $\times2$ \hfill $256{\times}40^2$};
\node[L] (n12) at (\xn,-10.1) {\textbf{12} Concat 11, $\tilde P_4$ \hfill $384{\times}40^2$};
\node[L] (n13) at (\xn,-9.0)  {\textbf{13} C3k2 \hfill $128{\times}40^2$};
\node[L] (n14) at (\xn,-7.9)  {\textbf{14} Upsample $\times2$ \hfill $128{\times}80^2$};
\node[L] (n15) at (\xn,-6.8)  {\textbf{15} Concat 14, $\tilde P_3$ \hfill $256{\times}80^2$};
\node[L] (n16) at (\xn,-5.7)  {\textbf{16} C3k2 \hfill $64{\times}80^2$};
\foreach \a/\b in {n11/n12,n12/n13,n13/n14,n14/n15,n15/n16}
  \draw[arr] (\a) -- (\b);
\def\xm{13.6}
\node[L] (n17) at (\xm,-5.7)  {\textbf{17} Conv $3{\times}3$ s2 \hfill $64{\times}40^2$};
\node[L] (n18) at (\xm,-6.8)  {\textbf{18} Concat 17, 13 \hfill $192{\times}40^2$};
\node[L] (n19) at (\xm,-7.9)  {\textbf{19} C3k2 \hfill $128{\times}40^2$};
\node[L] (n20) at (\xm,-9.0)  {\textbf{20} Conv $3{\times}3$ s2 \hfill $128{\times}20^2$};
\node[L] (n21) at (\xm,-10.1) {\textbf{21} Concat 20, $\tilde P_5$ \hfill $384{\times}20^2$};
\node[L] (n22) at (\xm,-11.2) {\textbf{22} C3k2 \hfill $256{\times}20^2$};
\foreach \a/\b in {n17/n18,n18/n19,n19/n20,n20/n21,n21/n22}
  \draw[arr] (\a) -- (\b);
\draw[arr] (n16) -- (n17);
\draw[arr] (n13.east) -- ++(0.15,0) |- (n18.west);
\draw[arr] (s5) -- (n11) node[pos=0.12, above] {$\tilde P_5$};
\draw[arr] (s4.east) -- (7.1,-6.9) node[pos=0.5, above] {$\tilde P_4$} |- (n12.west);
\draw[arr] (s3.east) -- (7.4,-4.4) node[pos=0.5, above] {$\tilde P_3$} |- (n15.west);
\draw[arr] (s5.south) -- ++(0,-0.45) -| ([xshift=-0.2cm]n21.west) -- (n21.west);
\def\xh{17.0}
\node[A, text width=2.1cm, minimum height=6.1cm, align=center] (att) at (\xh,-8.45)
  {\textbf{Cross-scale attention}\\[2pt] Eq.~(\ref{eq:attention})\\[2pt] channel MLP\\ $448{\rightarrow}56{\rightarrow}448$\\[2pt] spatial $7{\times}7$ conv\\ per scale\\[2pt] gains $\gamma_l = 0$\\ at start};
\draw[arr] (n16.north) -- ++(0,0.35) coordinate (n3t) -| ([xshift=-0.22cm]att.north west) |- ([yshift=2.4cm]att.west);
\node[above left, inner sep=1pt] at ([xshift=-0.22cm]att.north west |- n3t) {$N_3$};
\draw[arr] (n19.east) -- (n19.east -| att.west) node[pos=0.5, above] {$N_4$};
\draw[arr] (n22.east) -- (n22.east -| att.west) node[pos=0.5, above] {$N_5$};
\def\xp{20.55}
\node[L, text width=3.6cm, minimum height=2.3cm, align=left] (pose) at (\xp,-6.6)
  {\textbf{23} Pose head ($\tilde N_3$, $\tilde N_4$, $\tilde N_5$)\\[2pt] box: $4\times16$ DFL bins\\ classes: $LEFT$, $RIGHT$, $NEUTRAL$\\ keypoints: $17\times(x, y, v)$\\ 8{,}400 anchor points};
\node[A, text width=3.6cm, minimum height=1.6cm, align=left] (dist) at (\xp,-10.3)
  {\textbf{Distance head}\\[2pt] per scale: 2 Conv $3{\times}3$\\ (16 ch.) and a $1{\times}1$ conv\\ distance $\hat d$ per\\ anchor point};
\draw[arr] (att.east |- pose.west) -- (pose.west) node[pos=0.5, above] {$\tilde N_l$};
\draw[darr] (att.east |- dist.west) -- (dist.west);
\node[draw=black!40, dashed, rounded corners=1.5pt, align=left, text width=4.6cm, inner sep=3pt, anchor=north west] (loss) at (18.6,-0.15)
  {\textbf{Training loss} $\mathcal{L}_{total}$, Eq.~(\ref{eq:loss_weights}):\\
   pose head: $\mathcal{L}_{box}$ (CIoU, DFL), $\mathcal{L}_{conf} + \mathcal{L}_{gesture}$ (BCE), $\mathcal{L}_{kpt}$, $\mathcal{L}_{kobj}$\\
   distance head: $\mathcal{L}_{depth}$ (L1)\\
   gesture logits of 4-frame clips: $\mathcal{L}_{temporal}$};
\draw[arr, dashed, black!50] (pose.north) -- (pose.north |- loss.south);
\draw[arr, dashed, black!50] (dist.east) -| ([xshift=-0.3cm]loss.south east);
\node[sec] at (\xd, 1.75) {Depth stream};
\node[sec] at (\xb, 1.75) {Backbone};
\node[sec] at (11.5, -4.5) {Neck (FPN + PAN)};
\node[sec] at (18.7, -4.5) {Head};
\node[L, text width=0.45cm, minimum height=0.3cm] (lg1) at (9.0,0.6) {};
\node[anchor=west, align=left] at (9.35,0.6) {YOLOv11n-pose layer, unchanged (COCO weights):\\ layer number, type, output channels${\times}$height$^2$};
\node[A, text width=0.45cm, minimum height=0.3cm] (lg2) at (9.0,0.0) {};
\node[anchor=west, align=left] at (9.35,0.0) {added module (trained from scratch)};
\node[sum] at (9.0,-0.6) {$+$};
\node[anchor=west, align=left] at (9.35,-0.6) {element-wise addition, Eq.~(\ref{eq:fusion}), $W_l = 0$ at start};
\end{tikzpicture}%
}
\caption{Architecture of the modified YOLOv11 (YOLOv11-$\mathcal{L}_{total}$, Section~\ref{sec:t4_impl}). Grey: the unchanged YOLOv11n-pose layers 0--23 with their COCO weights (layer number, type and output size, channels$\times$height$^2$, for a $640\times640$ input). Blue: the added depth stream, its fusion into the backbone outputs $P_3$, $P_4$, $P_5$ after layers 4, 6 and 10 (Eq.~(\ref{eq:fusion})), the cross-scale attention (Eq.~(\ref{eq:attention})) and the distance head.}
\label{fig:yolo11arch}\label{fig:t4_arch}
\end{figure*}
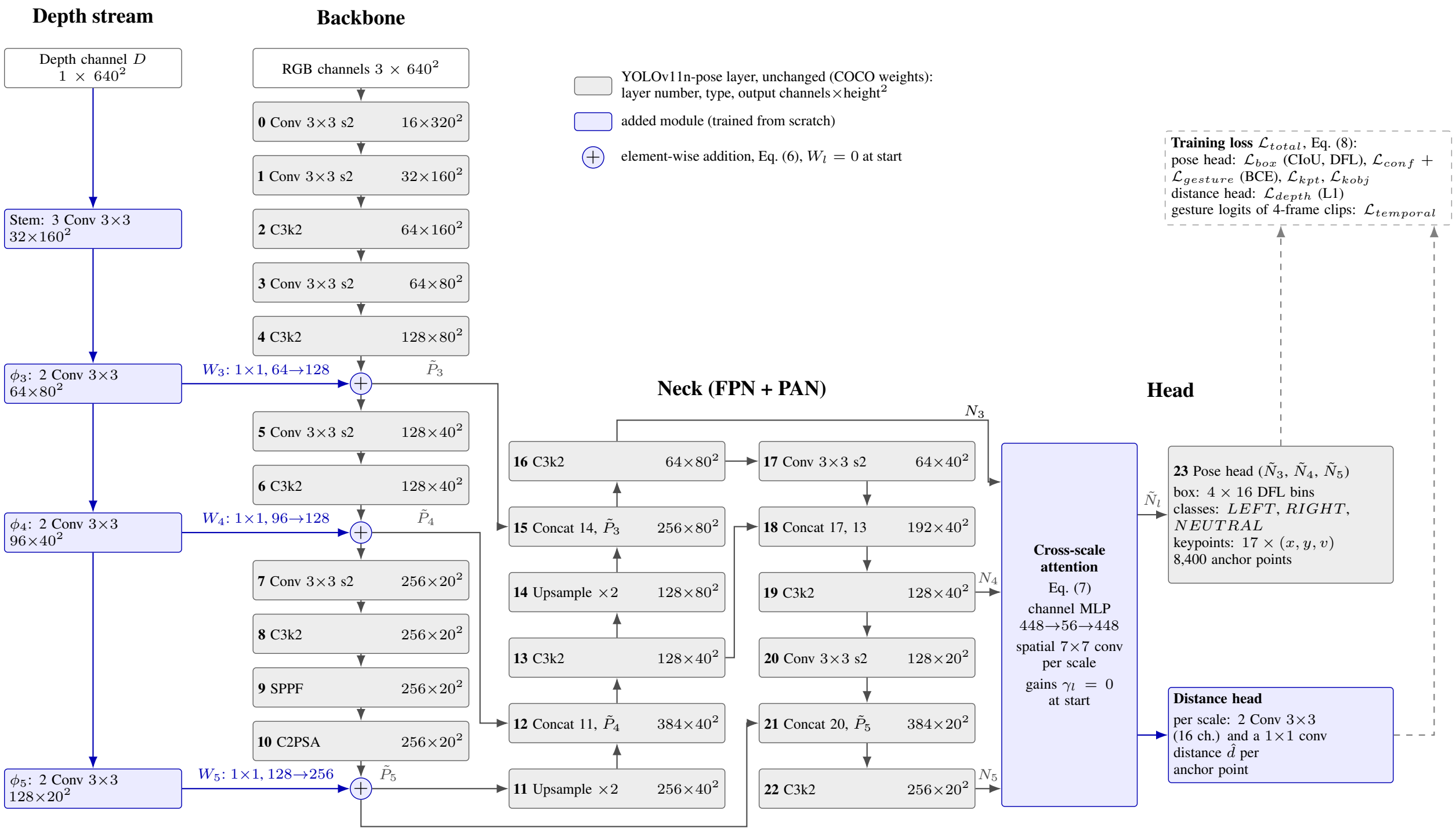

YOLO is a family of real-time detectors that predict all objects in a single pass. Our framework uses YOLOv11n-pose \cite{khanam_yolov11_2024}, extended to RGB-D input as YOLOv11-$\mathcal{L}_{total}$ (Fig.~\ref{fig:yolo11arch}, Section~\ref{sec:t4_impl}), with four parts:

{
\begin{itemize}
    \item The Depth stream, a parallel convolutional network, extracts depth features at strides 8, 16 and 32 for the depth-aware FPN.
    \item The Backbone (layers 0--10) computes the RGB features $P_3$, $P_4$ and $P_5$ with Conv, C3k2, SPPF and C2PSA blocks and fuses them with the depth features after layers 4, 6 and 10.
    \item The Neck (layers 11--22) combines the fused features across scales with an FPN and a PAN, and a cross-scale attention re-weights them for each frame.
    \item The Head (layer 23) predicts boxes, the gestures $LEFT$, $RIGHT$ and $NEUTRAL$ and 17 keypoints without anchors, and a distance head predicts each person's distance.
\end{itemize}
}

At the core of the recognition pipeline lies YOLOv11---a state-of-the-art, real-time object detection framework that we adapt for gesture classification and human pose estimation. By processing RGB frames, YOLOv11 detects the presence of a human operator and extracts spatial keypoints (e.g., shoulders, elbows, hands). Using these keypoints, the system classifies gestures such as left-hand pointing, right-hand pointing, or neutral stance, which are used as high-level commands to control the UAV's movement direction. YOLOv11's low latency and high accuracy ensure the UAV reacts swiftly and reliably to human input, which is vital in dynamic SAR operations \cite{alif_yolov11_2024}.
\subsection{Perception Node (ROS2)} The Perception Node handles the synchronisation and fusion of the RGB and depth data streams. Implemented in ROS2, this node utilises ROS2 launch mechanisms and image transport tools to perform time-stamped alignment of the camera feeds. Once aligned, the combined data is passed through a pre-processing pipeline that includes noise filtering, Region of Interest (ROI) extraction, and transformation into a format suitable for gesture classification. The output of this node forms the basis for higher-level processing in the Gesture Recognition Node.
\subsection{Gesture Recognition Node (YOLOv11-based)} This node houses the gesture inference model and acts as the decision-making layer. Once the depth-fused RGB-D data reaches this node, the YOLOv11 model is applied to detect body pose and interpret the gesture command. The node emits a symbolic command (e.g., ``MOVE LEFT'' or ``MOVE RIGHT'') based on the interpreted gesture. The commands are published as ROS2 messages on a dedicated topic, which is subscribed to by the Control Node.
\subsection{Control Node (ROS2)} The Control Node translates high-level gesture commands into low-level UAV control signals. It acts as the interface between the gesture logic and the UAV's flight controller. Implemented as a ROS2 node, it subscribes to gesture command topics, maps each command to predefined velocity or position setpoints, and publishes them to the PX4 flight stack via MAVROS \cite{ermakov_mavros_2025} or FastRTPS \cite{eprosima_eprosima_2025} protocols. Real-time behaviour is maintained using ROS2's DDS, which ensures bounded latency and deterministic delivery of control signals.
\subsection{PX4 Flight Stack} The PX4 Flight Stack is a widely used open-source flight control system that communicates directly with the UAV hardware. In our simulation-based setup, it receives control signals from the Control Node and executes corresponding UAV manoeuvres -- directional flight, hovering, or turning. PX4 is responsible for implementing the low-level dynamics, sensor integration, and actuator control necessary for flight stability and execution fidelity.
\subsection{Gazebo Simulator (SAR Environment Simulation)} To validate the system in a realistic SAR context, the entire framework is tested within the Gazebo simulator. Gazebo provides a physics-accurate environment in which terrain, obstacles, lighting, and sensor noise can be simulated to replicate real-world rescue conditions \cite{koenig_design_2004}. Within this environment, the UAV’s motion, gesture recognition, and system responses can be closely analysed, allowing for iterative debugging, performance benchmarking, and scenario-based testing.
\subsection{System Integration via ROS2 DDS Middleware} All modules communicate over ROS2 topics, nodes, and services using the DDS middleware. DDS ensures low latency, real-time message passing with Quality-of-Service (QoS) settings tailored for high-frequency, time-critical applications. This design choice ensures synchronisation between the sensing, decision, and actuation stages, which is vital in a time-sensitive domain such as search-and-rescue.

\section{Methodology}
\label{sec:meth_sec}

\subsection{Sensor Integration and Data Processing}
The proposed system uses multimodal sensor fusion with the Intel RealSense D435i, an RGB-D camera that provides high-resolution colour images and aligned depth maps. This dual-channel input significantly enhances the UAV's robustness and spatial understanding, which are critical for real-time decision-making in SAR scenarios.

\subsubsection{RGB-D and Depth Sensing} The RGB stream extracts the visual details needed to detect gestures and body keypoints, while the depth stream provides a 3D representation of the surrounding environment. This integration allows the UAV to estimate human gestures not only from 2D appearance but also from 3D positioning, even in low-light or visually cluttered environments. The depth map enhances gesture detection by separating the human subject from the background, reducing false positives due to occlusion or scene ambiguity.
\subsubsection{Depth-Weighted Human Pose Estimation} To achieve reliable gesture-driven control, we formulate a fused 3D pose estimation strategy. The estimated 3D joint position vector \(\vec{P}_j \in \mathbb{R}^3\) for each skeletal joint \(j\) is computed as a weighted fusion of multimodal cues:

\begin{equation}
\label{eq:multimodal}
\vec{P}_j = \alpha \cdot \vec{D}_j + \beta \cdot \vec{RGB}_j + \gamma \cdot \vec{T}_j + \delta \cdot \vec{C}_j
\end{equation}

\begin{itemize}
    \item $\vec{D}_j$ represents the depth-derived 3D skeletal keypoints for joint $j$.
    \item $\vec{RGB}_j$ is the colour-based 3D joint estimate obtained from RGB images using detectors like YOLOv11.
    \item $\vec{T}_j$ captures the temporal smoothing or motion consistency over time for joint $j$ to reduce jitter.
    \item $\vec{C}_j$ encodes contextual priors from the environment (e.g., ground plane constraints, obstacles) relevant to joint $j$.
\end{itemize}

The fusion weights $\alpha, \beta, \gamma, \delta$ are dynamically adjusted based on sensor reliability according to:

\begin{equation}
\label{eq:weights}
\alpha = \frac{e^{-\lambda_d \sigma_d^2}}{\sum\limits_{i \in \{d, \text{rgb}, t, c\}} e^{-\lambda_i \sigma_i^2}}
\end{equation}

\begin{center}
    \footnotesize \itshape $\alpha$ is for depth weight. Similar expressions apply for $\beta$, $\gamma$, and $\delta$, which correspond to the weights for $\vec{RGB}_j$, $\vec{T}_j$, and $\vec{C}_j$ respectively.
\end{center}

where $\sigma_d^2$ represents the variance in depth measurements, and $\lambda_d$  is a tunable confidence decay factor. This formulation allows the system to prioritise the most reliable modality under varying environmental conditions. For example, in low light, $\alpha$ becomes dominant, giving more weight to the depth modality.

This adaptive multimodal fusion enables accurate 3D pose estimation, thereby enhancing both gesture recognition accuracy and the spatial alignment of UAV responses. This is especially critical in SAR operations, where misinterpreted gestures or incorrect spatial judgements could result in mission failure or serious safety hazards.

\subsection{Real-Time Gesture Recognition with YOLOv11}

In the proposed SAR framework, real-time gesture recognition is a critical component that enables intuitive human-UAV collaboration. Unlike generic object detectors, our modified version of YOLOv11 \cite{khanam_yolov11_2024} incorporates depth-awareness, temporal modelling, and adaptive attention to address challenges such as occlusions, background noise, and rapid movement inherent to SAR scenarios. These enhancements are crucial, as UAV-captured footage often involves changing lighting conditions, complex backgrounds, and sudden camera movements. Incorporating depth data strengthens the model's capability to separate humans from background clutter, temporal modelling increases stability by analysing motion across frames, and the adaptive attention mechanism helps the network concentrate on important features while filtering out irrelevant noise.

\subsubsection{YOLOv11-Based Gesture Classification} YOLOv11 is a one-stage object detection network that simultaneously performs bounding box regression, object classification, and confidence estimation \cite{vijayakumar_yolo-based_2024, liu_object_2018}.
For our task, it is extended with the following key enhancements:

\begin{itemize}
\item[a)] \textit{Depth-Aware Feature Pyramid Network (FPN)}: Traditional FPNs operate on RGB features alone. Our modification adds a parallel depth stream to fuse RGB and depth features, improving detection of small and occluded gestures.

\item[b)] \textit{Anchor-Free Detection}: Predefined anchor boxes can struggle to capture the variability of hand sizes and gestures. To address this, anchor-free detectors eliminate anchors entirely and instead predict object centres and bounding box dimensions directly, which makes the model efficient and adaptable to our UAV gesture dataset.

\item[c)] \textit{Multi-Scale Attention Mechanisms}: We incorporate cross-scale attention blocks that learn to emphasise salient regions (e.g., hands or arm joints) while suppressing background distractions. This mechanism is especially helpful in SAR scenes where gesture articulation may vary due to clothing or motion blur.
\end{itemize}

\subsubsection{Custom Multi-Task Loss Function} The learning objective is formulated as a weighted combination of spatial, semantic, and temporal losses. The spatial loss measures how well the model predicts the correct spatial locations of keypoints within each frame. The semantic loss evaluates the model's ability to capture the correct gesture class. The temporal loss assesses the consistency and smoothness of predictions across consecutive frames :

\begin{equation}
\label{eq:loss_fun}
\begin{aligned}
\mathcal{L}_{total} ={}& \mathcal{L}_{box} + \mathcal{L}_{conf} + \mathcal{L}_{depth} \\ & + \mathcal{L}_{gesture} + \mathcal{L}_{temporal}
\end{aligned}
\end{equation}

\begin{itemize}
    \item $\mathcal{L}_{box}$, $\mathcal{L}_{depth}$, $\mathcal{L}_{conf}$ are spatial losses whereas $\mathcal{L}_{gesture}$ is semantic  and $\mathcal{L}_{temporal}$ is temporal loss.
    \item $\mathcal{L}_{box}$: IoU loss and DFL loss for precise bounding box regression .
    \item $\mathcal{L}_{conf}$: Binary cross-entropy loss penalising misclassification of presence or absence of gesture regions.
    \item $\mathcal{L}_{gesture}$: Binary cross-entropy loss of the gesture classes ($LEFT$, $RIGHT$ and $NEUTRAL$).
    \item $\mathcal{L}_{depth}$: Enforces geometric coherence between RGB-based detection and the corresponding depth map using $\mathcal{L}_{1}$ loss.
    
    \item $\mathcal{L}_{temporal}$: A temporal consistency term that minimises variations in gesture logits across consecutive frames, encouraging stable predictions over time.
\end{itemize}

\section{Simulation Environment}
\label{sec:sim_sec}

\subsection{Human Model Generation and Integration in Gazebo}
We have created a custom human mesh for realistic human-robot interaction within the Gazebo environment by utilising Blender, a widely used open-source 3D modelling tool, to design the human model \cite{community_blender_2018}. The base human anatomy was sculpted and rigged using Blender's armature system, allowing for joint manipulation and animation playback. This rigging was essential for testing dynamic human gestures, such as left/right pointing, which are critical for interaction studies.

The final rigged model was exported in the COLLADA (\textit{.dae}) format, which preserves the mesh geometry and skeleton information. This \textit{.dae} file was then embedded within the Simulation Description Format (SDF) file to allow seamless import into Gazebo. The SDF specification includes references to the model's mesh, joint limits, and pose initialisation of the human model, ensuring correct visualisation and physics simulation within the virtual environment.

However, difficulties arose when integrating textures and clothing. To improve realism, we tried outfitting the human model with layered clothing meshes and applying external texture maps. These additional elements, however, introduced multiple dependencies during the COLLADA export. While \textit{.dae} supports such linked assets, Gazebo's SDF parser does not automatically resolve external file dependencies unless explicitly declared and co-located. This created a barrier for directly importing the clothed model into SDF, since Blender bundles these as relative paths that are not preserved during the model upload or packaging process.
Moreover, Gazebo does not provide a direct utility to convert SDF files back to COLLADA (.dae). This one-way pipeline means any post-rigging enhancements (like texture mapping or re-targeting skeletons) must be finalised within Blender before exporting. To mitigate this, we opted to use a simplified mesh (without clothing) in our Gazebo simulation while retaining realistic proportions and gestures for evaluation. In the future, we could include a unified mesh-texture baking and automated dependency packaging strategy to overcome this constraint.

\begin{algorithm}[]
\fontsize{6pt}{5pt}\selectfont
\caption{\enskip Unified-PDD Algorithm}
\label{alg:algo}
\begin{algorithmic}[1]
\Require Video stream (RGB or Depth), YOLOv11 pose model
\Ensure Annotated frames with distances, keypoints, and direction \Statex \hskip2em labels; high-level UAV commands

\State Initialise ROS2 node and subscribe to rgb\_camera and depth\_camera topic
\State Load YOLOv11 pose estimation model

\While{new frame is received}
    \If{frame contains depth data}
        \Statex \hskip4em Convert to OpenCV ($32FC1$), replace NaNs, clip to
        \Statex \hskip2em $[near, far]$
        \Statex \hskip4em Normalise and apply colourmap for pseudo-RGB input
    \Else
        \Statex \hskip4em Convert to OpenCV ($BGR8$) format
    \EndIf

    \vskip0.5em
    \State Run pose estimation on the input frame

    \If{detections exist}
        \ForAll{detected persons}
            \Statex \hskip4em Extract bounding box coordinates
            \If{depth data available}
                \Statex \hskip7em Get depth ROI and filter invalid values
                \Statex \hskip7em Compute median depth as estimated distance
            \Else~{Set distance as unknown or default far clipping value}
            \EndIf

            \vskip0.5em
            \State Extract 2D keypoints coordinates for pose, \Statex \hskip5em sides as seen by the UAV:
            \Statex \hskip7em If left/right keypoints appear mirrored, swap them
            \State Compute joint angles using keypoints:
            \Statex \hskip7em Elbow: $\angle$~(shoulder, elbow, wrist)
            \Statex \hskip7em Shoulder: $\angle$~(hip, shoulder, elbow)
            \vskip0.5em
            \State Classify pointing direction using joint angle thresholds:
            \Statex \hskip7em Arm extended $\Leftrightarrow$ shoulder $\in$ [65$^\circ$,106$^\circ$] $\wedge$ elbow $\in$ [100$^\circ$,180$^\circ$]
            \Statex \hskip7em Arm detected $\Leftrightarrow$ shoulder and elbow keypoint confidence $\geq 0.5$
            \Statex \hskip7em If both arms extended: direction $\gets$ \texttt{NEUTRAL}
            \Statex \hskip7em Else if one arm extended $\wedge$ other arm detected:
            \Statex \hskip9em direction $\gets$ \texttt{RIGHT} (right arm) or \texttt{LEFT} (left arm)
            \Statex \hskip7em Else:  direction $\gets$ \texttt{NEUTRAL}
            \vskip0.5em
            \State{Map gesture to high-level UAV command:}
            \Statex \hskip7em \texttt{RIGHT} $\rightarrow$ Turn or move UAV left
            \Statex \hskip7em \texttt{LEFT} $\rightarrow$ Turn or move UAV right
            \Statex \hskip7em \texttt{NEUTRAL} $\rightarrow$ Hold or hover
            \vskip0.5em
            \State Annotate image with:
            \Statex \hskip7em Bounding box and estimated distance
            \Statex \hskip7em Pose keypoints and skeleton
            \Statex \hskip7em Direction label
        \EndFor
    \Else
        \State Display input image with message ``No person detected''
    \EndIf

    \Statex \hskip4em Display annotated frame
    \EndWhile
\end{algorithmic}
\end{algorithm}

\subsection{Algorithm}
We propose Unified-PDD, an algorithm for Pose estimation, Distance calculation, and Direction classification from UAV data, as shown in Algorithm.~\ref{alg:algo}, which performs real-time human pose estimation, distance measurement, and gesture-based direction recognition using RGB or depth video input. Each incoming frame is preprocessed according to its modality, i.e., if the incoming frame contains depth data, it is first converted into the OpenCV $32FC1$ format, which represents a 32-bit floating-point, single-channel image suitable for storing precise depth values.

Subsequently, any invalid or missing values -- caused by sensor noise, occlusions, or reflective surfaces -- are replaced with fallback values (e.g., $background$, $far$, $near$ or the $mean\_depth$) to ensure numerical stability. The resulting depth values are then clipped to a valid range 
$[near, far]$, corresponding to the minimum and maximum sensing limits of the depth camera, which are predefined and ensure all pixel values fall within a physically meaningful range. After clipping, the depth map is normalised and a colour map is applied, producing a pseudo-RGB image that encodes depth using colour gradients. This transformation enables models expecting 3-channel input (YOLO) to process depth data similarly to RGB images. Conversely, if the frame contains RGB data instead of depth, it is directly converted into the OpenCV $BGR8$ format, the standard 8-bit, 3-channel image representation used in OpenCV for colour images.

After pre-processing the input frame, the algorithm utilises YOLOv11 for real-time human detection and pose estimation to identify individuals and extract their body keypoints. For each detected person during flight, the algorithm determines the bounding box coordinates to localise the individual within the frame. In case of availability of depth data, the corresponding ROI is extracted from the depth map, and the median depth within the ROI is computed to estimate the person's distance from the UAV. If depth data is not available, the distance is set to an unknown state or the default far clipping value.
Then the algorithm extracts 2D keypoint coordinates representing major body joints. Since the operator faces the UAV, their right shoulder should appear on the left of the image. If YOLOv11 mirrors the left/right labels (e.g., in depth images, which show no face), the left/right keypoints are swapped back, so that left and right are always as seen by the UAV. These keypoints are used to calculate relevant joint angles (e.g., at the elbow and shoulders), which are then employed to classify the person's pointing direction. Based on the defined thresholds, if the right shoulder angle lies within $[65^\circ, 106^\circ]$ and the right elbow angle within $[100^\circ, 180^\circ]$, the direction is classified as $RIGHT$; if the left shoulder angle lies within $[65^\circ, 106^\circ]$ and the left elbow angle within $[100^\circ, 180^\circ]$, the direction is classified as $LEFT$; otherwise, the direction defaults to $NEUTRAL$. An arm meeting both thresholds counts as extended, and an arm counts as detected if its shoulder and elbow keypoint confidence is at least 0.5. If both arms are extended, the direction is $NEUTRAL$; if one arm is extended and the other is detected, the extended arm gives the direction; otherwise (e.g., the other arm hidden by fog), it is $NEUTRAL$.

This directional intent is used to issue high-level navigational commands to the UAV (e.g., turning, aligning, or holding position). As the operator faces the UAV, a $RIGHT$ gesture points to the UAV's left, so the UAV turns left for $RIGHT$, right for $LEFT$, and hovers for $NEUTRAL$.
Finally, the processed frame is annotated with the bounding boxes, detected keypoints, estimated distances, and direction labels, providing clear and interpretable visual feedback to facilitate gesture-based UAV control.

We used the pre-trained YOLOv11n-pose model (COCO weights) with the default settings of the official YOLOv11 implementation, without retraining. The retrained YOLOv11-$\mathcal{L}_{total}$ and its hyperparameters are described in Section~\ref{sec:training}.

\subsection{Simulation Environments for Gesture-Based UAV Control}

To evaluate the proposed UAV gesture interaction framework across varied visual and spatial conditions, we constructed three distinct Gazebo environments using publicly available and custom assets.

\begin{itemize}
\item{\textbf{Forest Environment:}} This scenario replicates a natural woodland setting with dense arrangements of palm and oak trees. The human model, generated via Blender, is positioned centrally and surrounded by trees in all directions. This setup simulates challenging natural conditions for visual gesture detection amidst foliage occlusion and irregular lighting as shown in Fig~\ref{fig:forest_env}.

\begin{figure}[H]
\centerline{\includegraphics[width=0.8\colfigwidth]{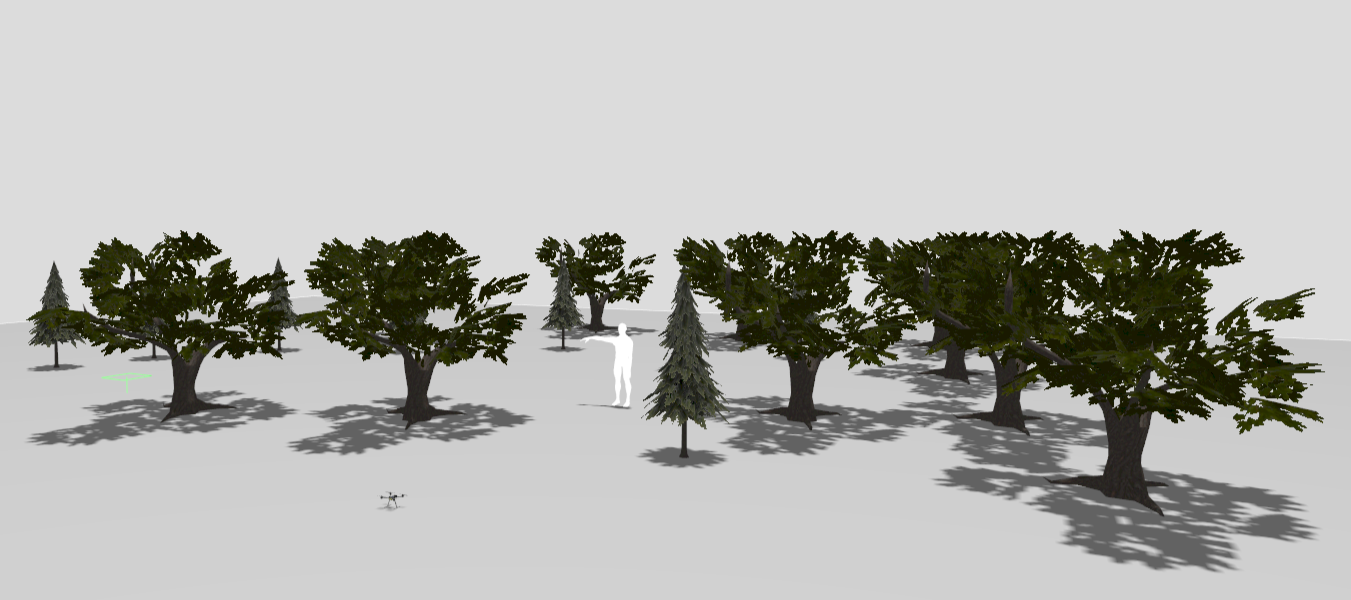}}
\caption{Forest Environment\label{fig:forest_env}}
\end{figure}

\item{\textbf{City Environment:}} Designed to represent an urban landscape, this environment includes structured elements such as buildings(including ruins of buildings), houses, a school, park rides, and trees. The human model is embedded within a typical city park-like area, providing a realistic interaction zone with urban clutter and background noise as shown in Fig~\ref{fig:city_env}.

\begin{figure}[H]
\centerline{\includegraphics[width=0.8\colfigwidth]{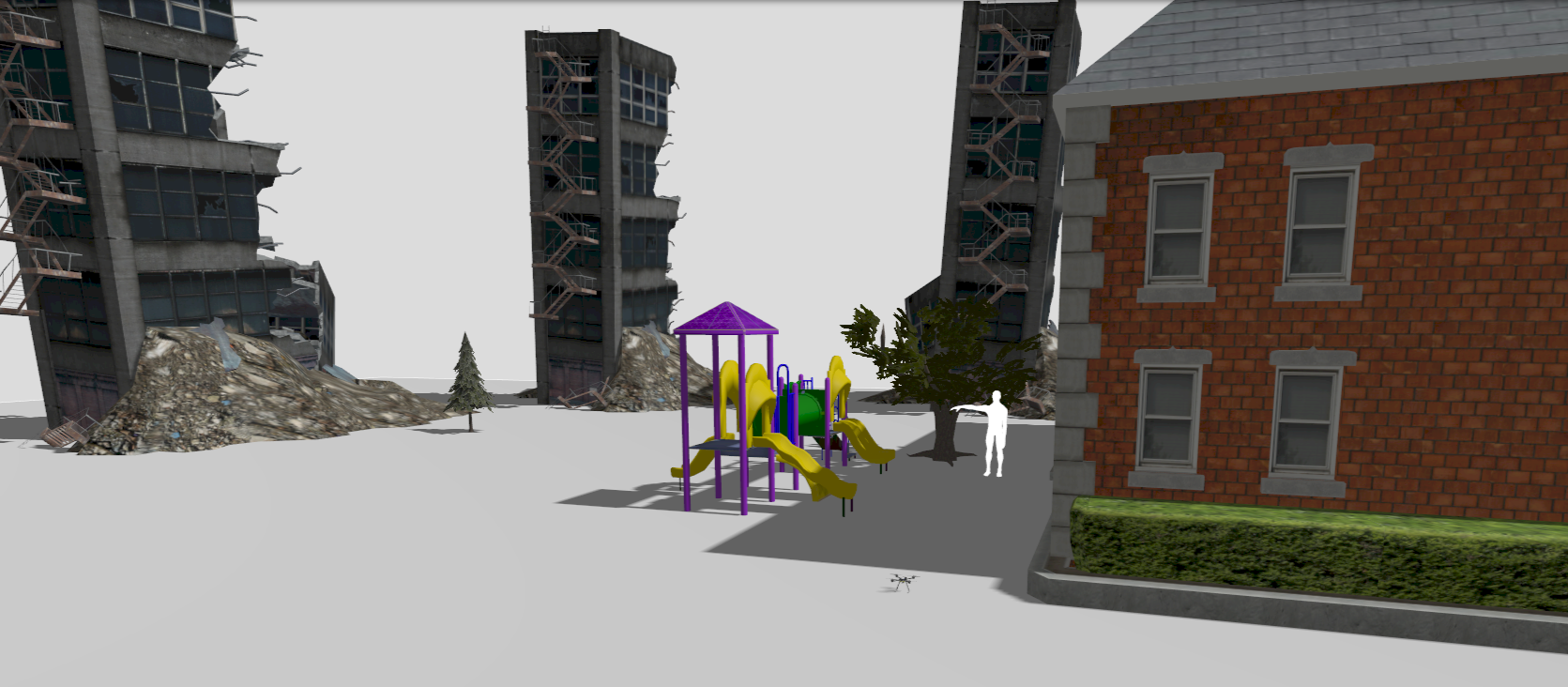}}
\caption{City Environment\label{fig:city_env}}
\end{figure}

\item{\textbf{Wall Environment:}} This minimalistic setup consists of a completely enclosed space using a high black wall. It removes all ambient visual complexity, isolating the human figure for controlled gesture recognition benchmarking as shown in Fig~\ref{fig:wall_env}.

\begin{figure}[H]
\centerline{\includegraphics[width=0.8\colfigwidth, height=0.4\colfigwidth]{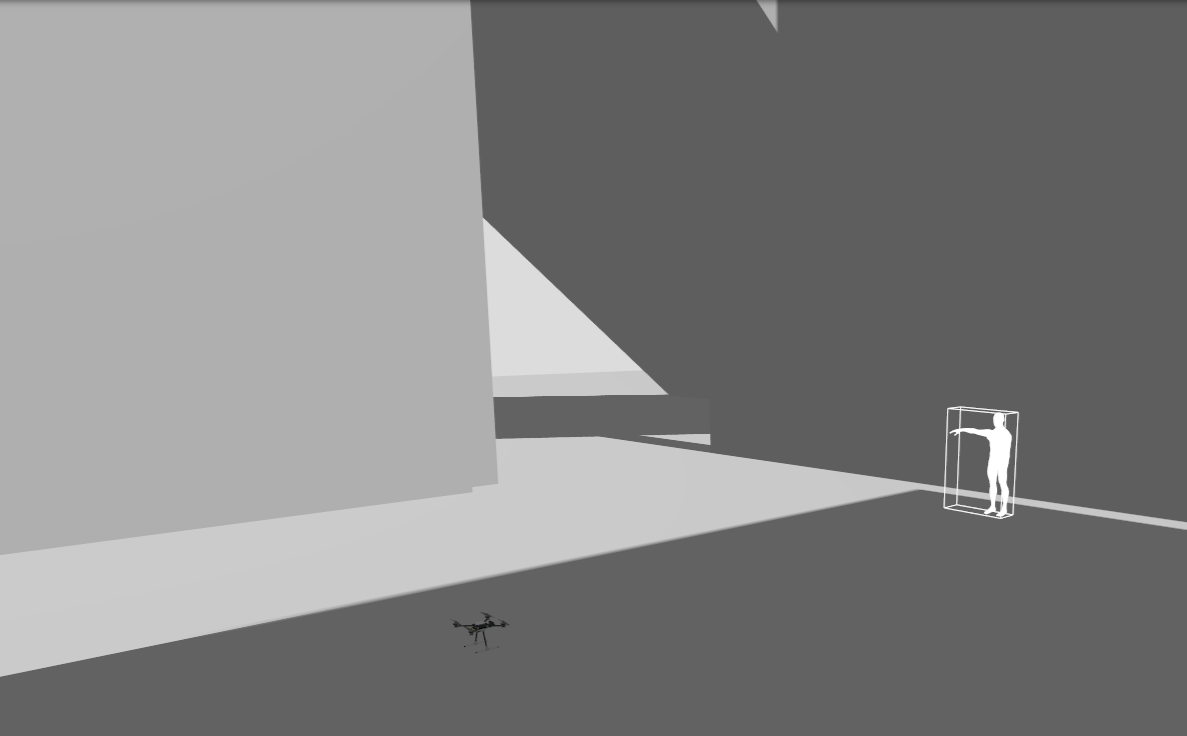}}
\caption{Wall Environment\label{fig:wall_env}}
\end{figure}
\end{itemize}

\paragraph*{Quantification of Visual Environment Complexity}:
To measure the complexity of different environments, we used \textit{Scene Entropy}, which quantifies the amount of information or complexity in a scene, on our custom environments. High scene entropy could mean the scene is very complex, detailed, and textured. Low scene entropy means the scene is less complex or homogeneous. This is done using Shannon Entropy \cite{shannon_mathematical_1948}, which is a measure of the uncertainty or complexity present in an image, reflecting the unpredictability of pixel intensity values or other visual features across the scene. There are two primary forms of entropy commonly considered -- \textit{global entropy} and \textit{local entropy}. Global entropy measures the overall uncertainty or complexity of the entire image, whereas local entropy captures spatial variations and fine details of the image, such as textures, edges, or noise. Considering global and local entropy facilitates a more detailed and accurate analysis by accounting for both the image's overall information content and the spatial variation of its features. Mathematically, for a discrete set of possible pixel intensity values with associated probabilities, the Shannon entropy \(H\) of an image is defined as:
    
\begin{equation}
\label{eq:shannon_entropy}
H_{scene} = - \sum_{i=1}^n P(x_i) \log_2 P(x_i)
\end{equation}

\begin{itemize}
    \item \(H_{scene}\) represents the Shannon entropy of the scene.
    \item \(P(x_i)\) represents the probability of occurrence of the intensity level \(x_i\), calculated as the ratio of the number of pixels with intensity \(x_i\) to the total number of pixels in the image.
    \item \(x_i\) is the \(i\)-th of the \(n\) pixel intensity values, where in an 8-bit greyscale image, \(x_i \in [0, 255]\) and \(n = 256\).
    \item The range of \(H\) for 8-bit greyscale image: (\( 0 \leq H \leq 8 \)).
\end{itemize}
    
\begin{table}[ht]
\centering
\caption{Scene Entropy}
\label{tab:scene_entropy}
\begin{tabular}{lcc}
\toprule
Environment & Global Entropy & Local Entropy \\
\midrule
Forest & 3.9928 & 1.6510 \\
City & 5.6586 & 2.8883 \\
Wall & 2.6269 & 0.3007 \\
\bottomrule
\end{tabular}
\end{table}

We can interpret from the Table~\ref{tab:scene_entropy} that for complex environments like Forest and City, the scene entropy is more than the scene entropy of a simple environment like Wall.

In addition to these, we also created fog-augmented variants of all three environments by enabling Gazebo's atmospheric fog plugin, thereby simulating reduced visibility conditions. This allowed assessment of gesture recognition robustness under varying weather conditions. All elements used are sourced from Gazebo-compatible assets, ensuring high fidelity and real-time rendering consistency during UAV simulation runs.

\begin{table}[t]
\centering
\caption{Skeleton Joints}
\label{tab:joints}
\begin{tabular}{|c|c||c|c|}
\hline
\textbf{Joints} & \textbf{Index} & \textbf{Joints} & \textbf{Index} \\
\hline
Nose & 0 & Left Wrist & 9 \\
Left Eye & 1 & Right Wrist & 10 \\
Right Eye & 2 & Left Hip & 11 \\
Left Ear & 3 & Right Hip & 12 \\
Right Ear & 4 & Left Knee & 13 \\
Left Shoulder & 5 & Right Knee & 14 \\
Right Shoulder & 6 & Left Ankle & 15 \\
Left Elbow & 7 & Right Ankle & 16 \\
Right Elbow & 8 &  &  \\
\hline
\end{tabular}
\end{table}

\section{Results and Analysis}
\label{sec:result_sec}

In this section, we present a comprehensive evaluation of our HITL gesture‐controlled UAV framework across three representative environments --- City, Forest, and Wall --- as well as under foggy conditions applied to each environment. All experiments were conducted in Gazebo (gz) with PX4 on a Ubuntu-based system (22.04 LTS), using the fog‐free versions of each environment (dense‐fog variants are evaluated separately). We focus on three core aspects of the framework:

\begin{enumerate}[I.]\item Keypoint Detection \item  Closest Correct Gesture Estimation \item Depth Estimation
\end{enumerate}

For each aspect, we focus on \textit{confidence} as the primary performance metric because our framework aims to enable reliable HITL control of UAVs through gesture recognition. Confidence scores provide a direct measure of the model's certainty in identifying relevant gestures, enabling rapid and reliable decision-making critical for real-time operations. Additionally, retraining the YOLO model on task data can achieve higher accuracy, further strengthening the reliability of confidence-based detections (Section~\ref{sec:ext_eval}). Table~\ref{tab:joints} shows the skeletal joints used for keypoint detection and CCE estimation.

We noted down the confidence scores manually during UAV flight time, and to ensure robustness, we repeated the process 10 times and reported the averaged values.

Section~\ref{sec:ext_eval} extends this evaluation to the modified YOLOv11, the depth-weighted pose estimation, a fixed simulated test set and real people.

\subsection{Keypoint Detection}

Beyond bounding‐box localisation, accurate detection of specific hand and arm keypoints is crucial for reliable gesture interpretation. The detection process does not take into account the distance, as it depicts the initial point when the keypoints are visible correctly. We evaluated YOLOv11's keypoint confidence \(\mathit{p}\) for RGB and depth perspective in Table~\ref{tab:keypoint_det_cce}.

\begin{table}[t]
    \centering
    \caption{Keypoint Detection and CCE Estimation}
    \label{tab:keypoint_det_cce}
    \begin{tabular}{c c c c c}
        \toprule
        \multirow{3}{*}{\textbf{Scenario}} & 
        \multicolumn{2}{c}{\textbf{Keypoint Detection}} & 
        \multicolumn{2}{c}{\textbf{Closest Correct Estimation}} \\
        \cmidrule(lr){2-3} \cmidrule(lr){4-5}
        & \textit{RGB} & \textit{Depth} & \textit{RGB} & \textit{Depth} \\
        & \(\mathit{p}\) & \(\mathit{p}\) & \(\mathit{p}\) & \(\mathit{p}\ \hspace{1cm}\mathit{d}\) \\
        \midrule
        \textbf{City}   & 0.79 & 0.34 & 0.83 & 0.87 \hspace{0.5cm} 7.13 \\
        \textbf{Forest} & 0.83 & 0.34 & 0.74 & 0.89 \hspace{0.5cm} 9.79 \\
        \textbf{Wall}   & 0.86 & 0.59 & 0.75 & 0.74 \hspace{0.5cm} 7.74 \\
        \bottomrule
    \end{tabular}
\end{table}

\subsubsection{\textbf{RGB‐Only Baseline}} Using only colour features, YOLOv11 localises model keypoints with a confidence of \SI{0.79}{} to \SI{0.86}{} across all scenes (Fig.~\ref {fig:keypoint_detection}), reflecting strong generalisation to diverse backgrounds. In contrast, depth cameras often provide noisy or inaccurate measurements because depth data lacks fine-grained detail which is must for keypoint detection.

\subsubsection{\textbf{Depth Baseline}} 
\begin{figure*}[tp]
    \centering
    \subfloat[\textit{(a)}]{\includegraphics[width=0.3\textwidth]{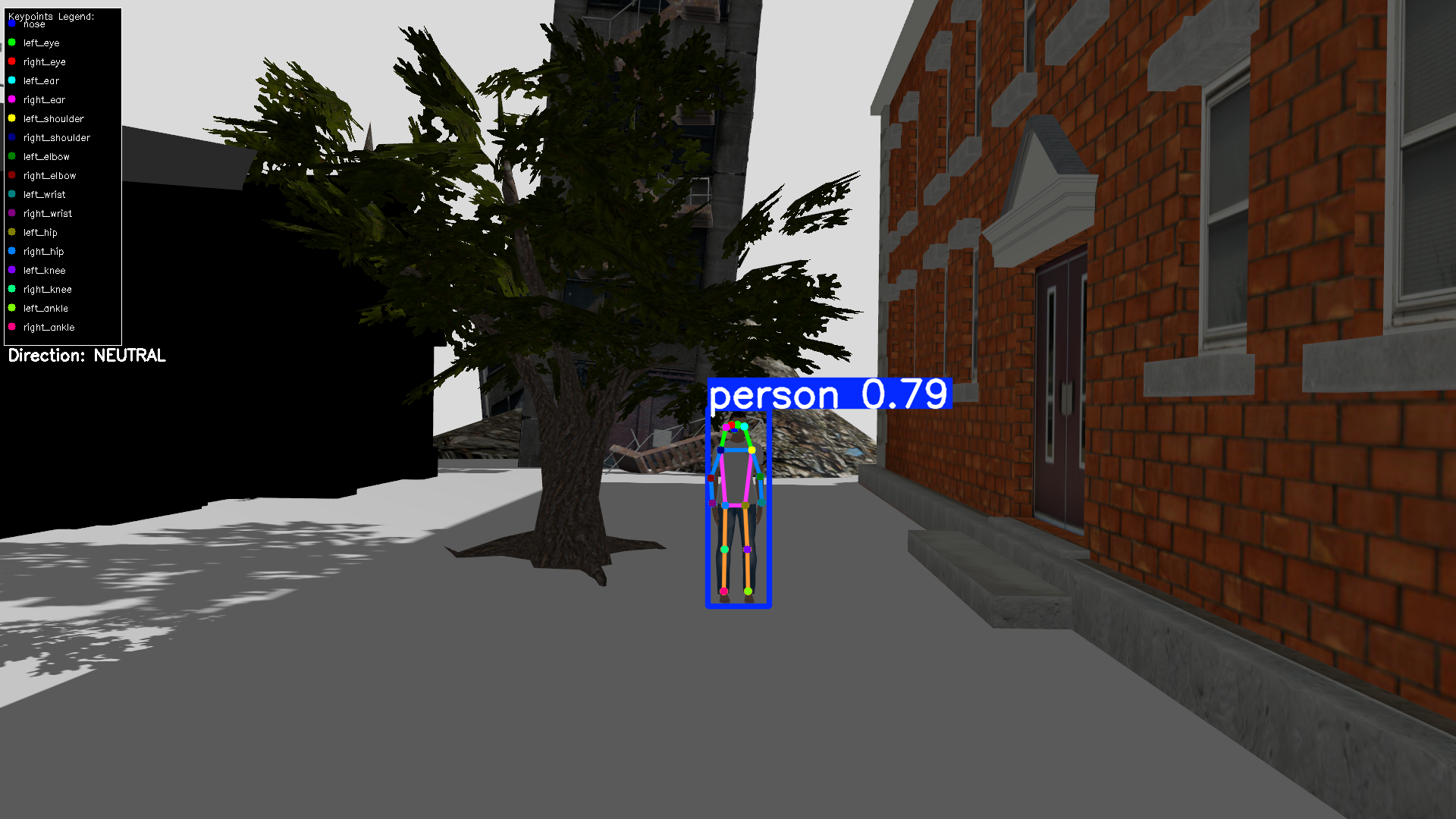}}\hfill
    \subfloat[\textit{(b)}]{\includegraphics[width=0.3\textwidth]{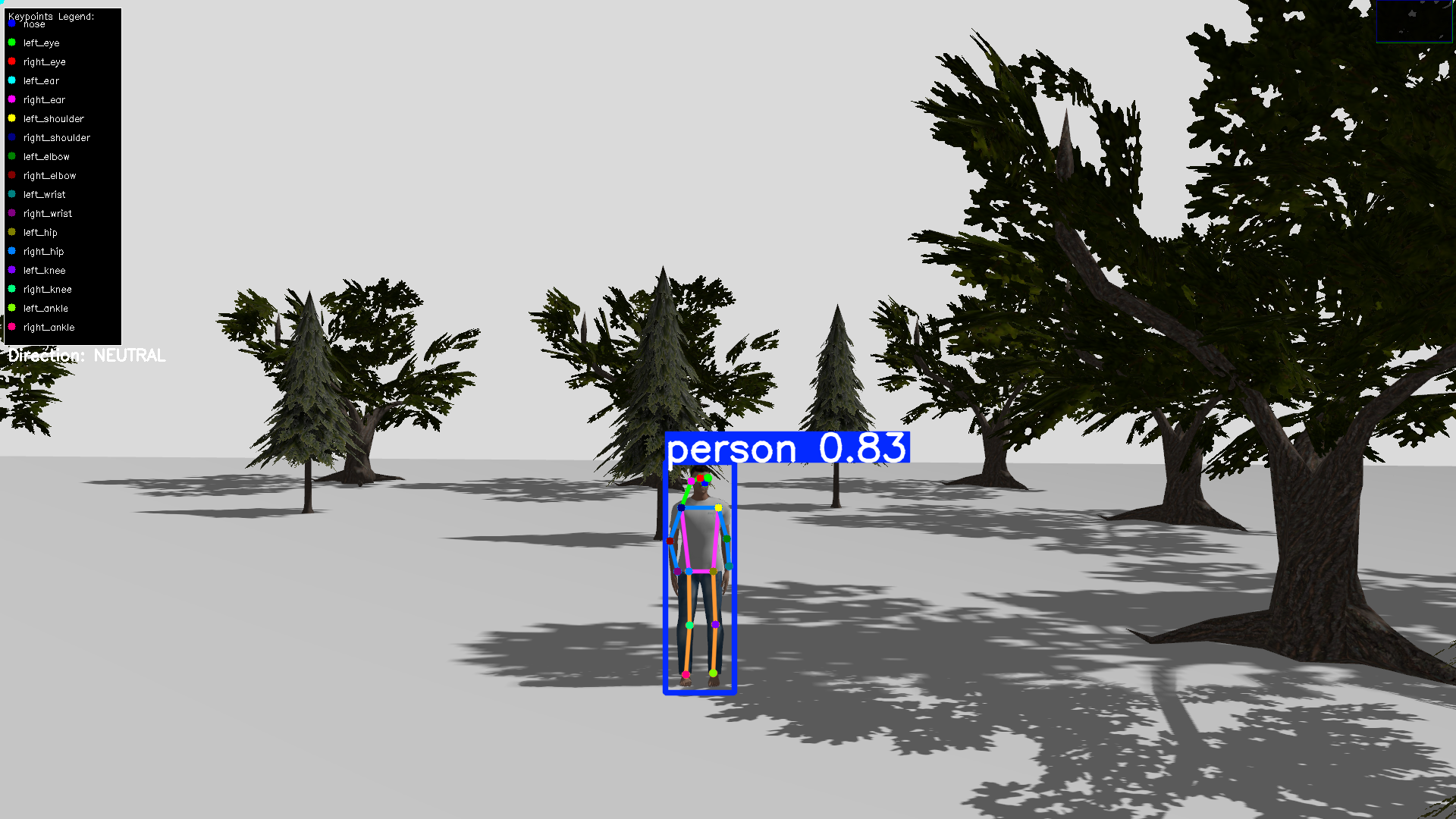}}\hfill
    \subfloat[\textit{(c)}]{\includegraphics[width=0.3\textwidth]{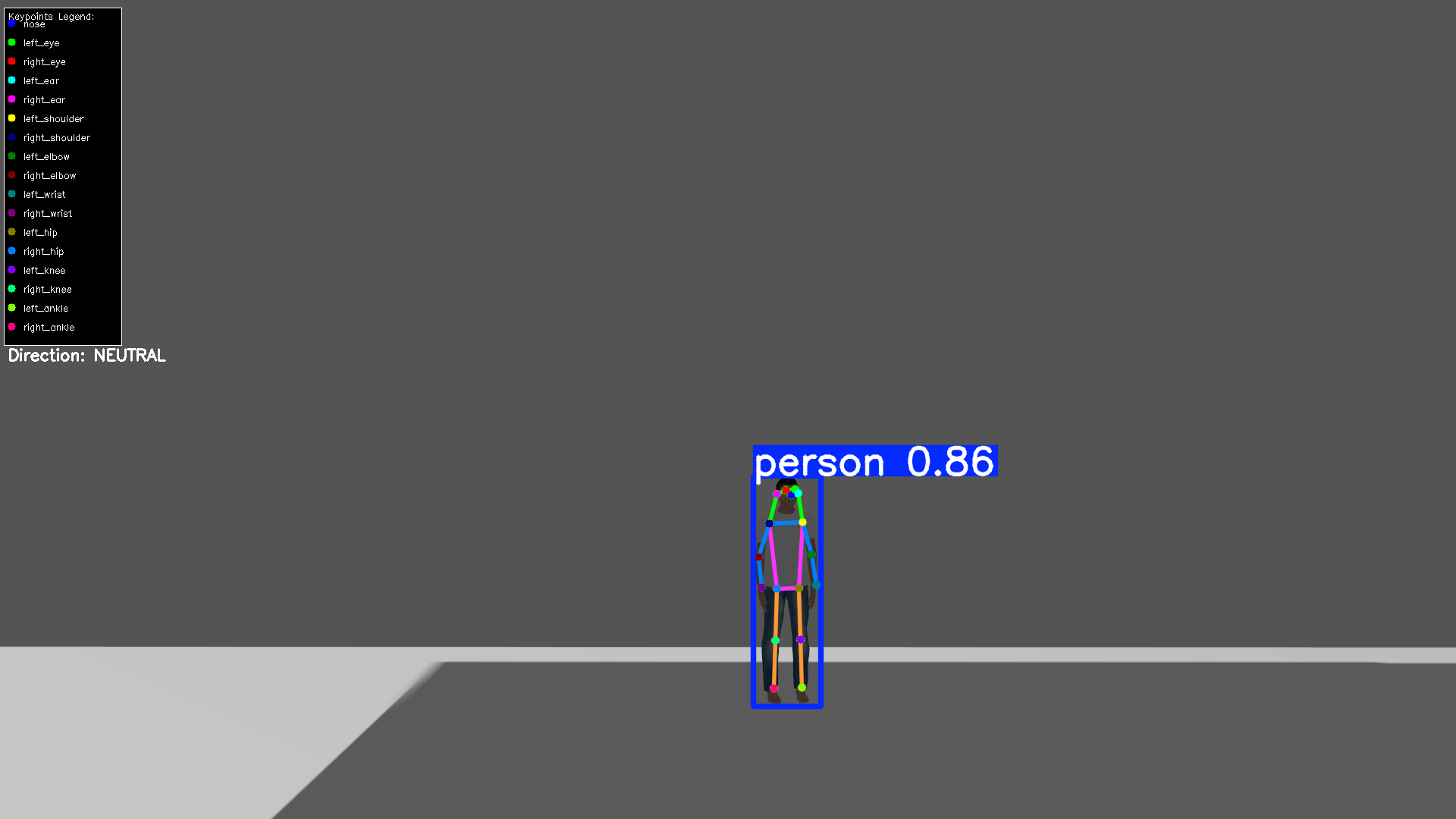}}\\
    \subfloat[\textit{(d)}]{\includegraphics[width=0.3\textwidth,height=0.18\textwidth]{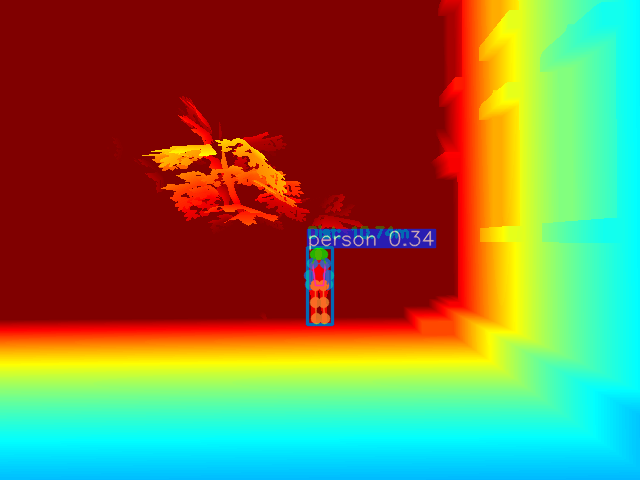}}\hfill
    \subfloat[\textit{(e)}]{\includegraphics[width=0.3\textwidth,height=0.18\textwidth]{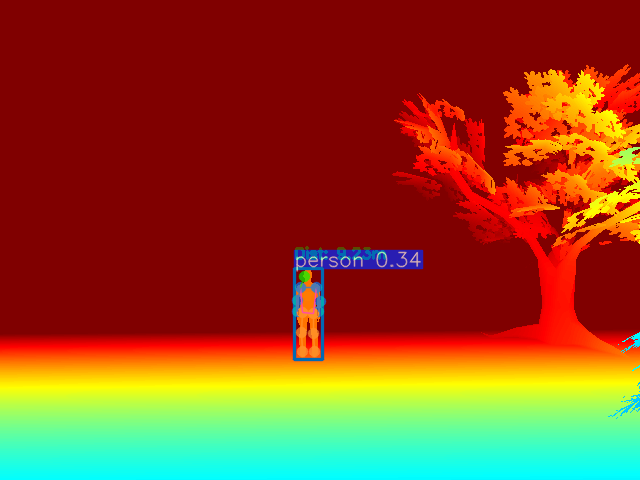}}\hfill
    \subfloat[\textit{(f)}]{\includegraphics[width=0.3\textwidth,height=0.18\textwidth]{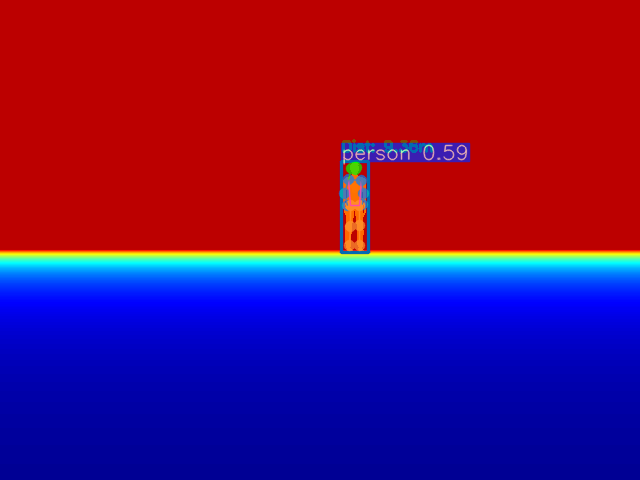}}\\
    \caption{(\textit{a,b,c}) are the RGB versions and (\textit{d,e,f}) are the Depth versions for City, Forest, and Wall scenarios, respectively, depicting the keypoint detection.}
    \label{fig:keypoint_detection}
\end{figure*}

Using depth features, YOLOv11 localises model keypoints with a confidence of \SI{0.34}{} to \SI{0.59}{} across all scenes (Fig.~\ref{fig:keypoint_detection}), indicating that, even at equal sensing distance, depth cameras lack the fine-grained spatial and texture cues available in RGB, making them less effective for accurate keypoint localisation on small or articulated structures like hands and arms.

These results indicate that depth does not measurably enhance keypoint detection robustness, due to lower spatial resolution, quantisation noise, and lack of texture cues. However, depth data can still be valuable for background segmentation and occlusion handling in scenarios where isolating the target from clutter or reconstructing a coarse 3D posture is more important than precise joint localisation.

\subsection{Closest Correct Gesture Estimation}
\begin{figure*}[tp]
    \centering
    \subfloat[\textit{(a)}]{\includegraphics[width=0.3\textwidth]{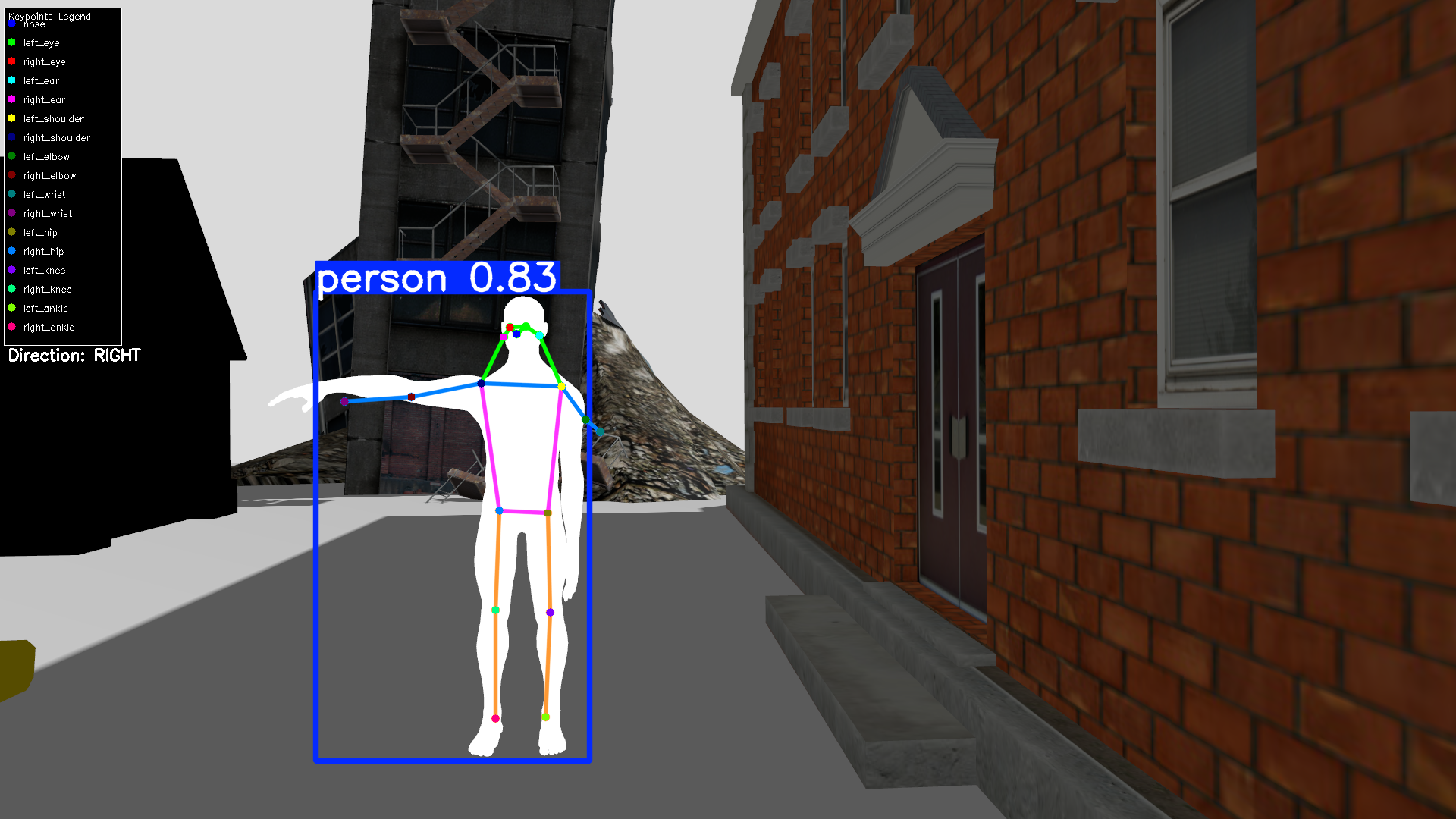}}\hfill
    \subfloat[\textit{(b)}]{\includegraphics[width=0.3\textwidth]{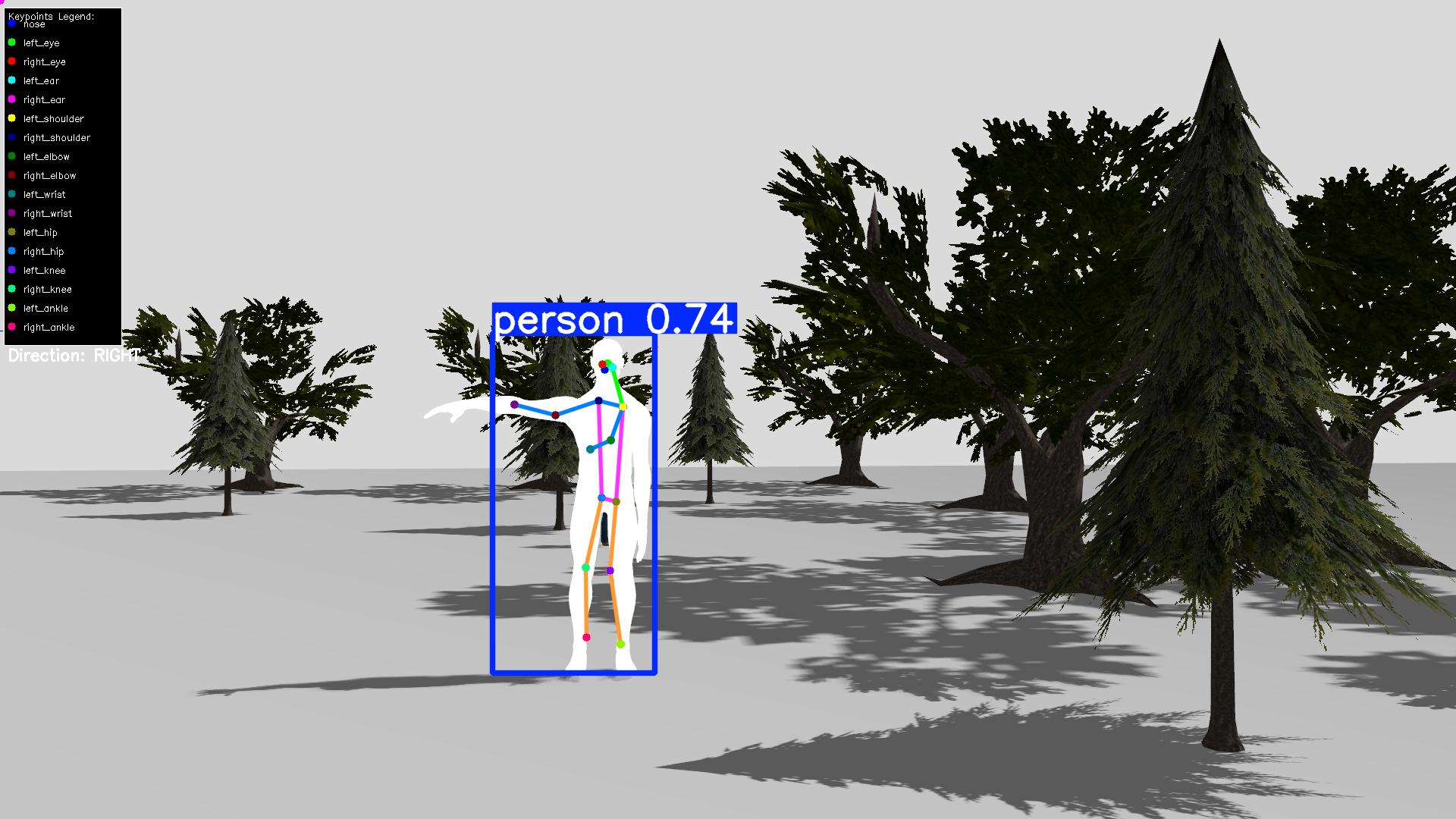}}\hfill
    \subfloat[\textit{(c)}]{\includegraphics[width=0.3\textwidth]{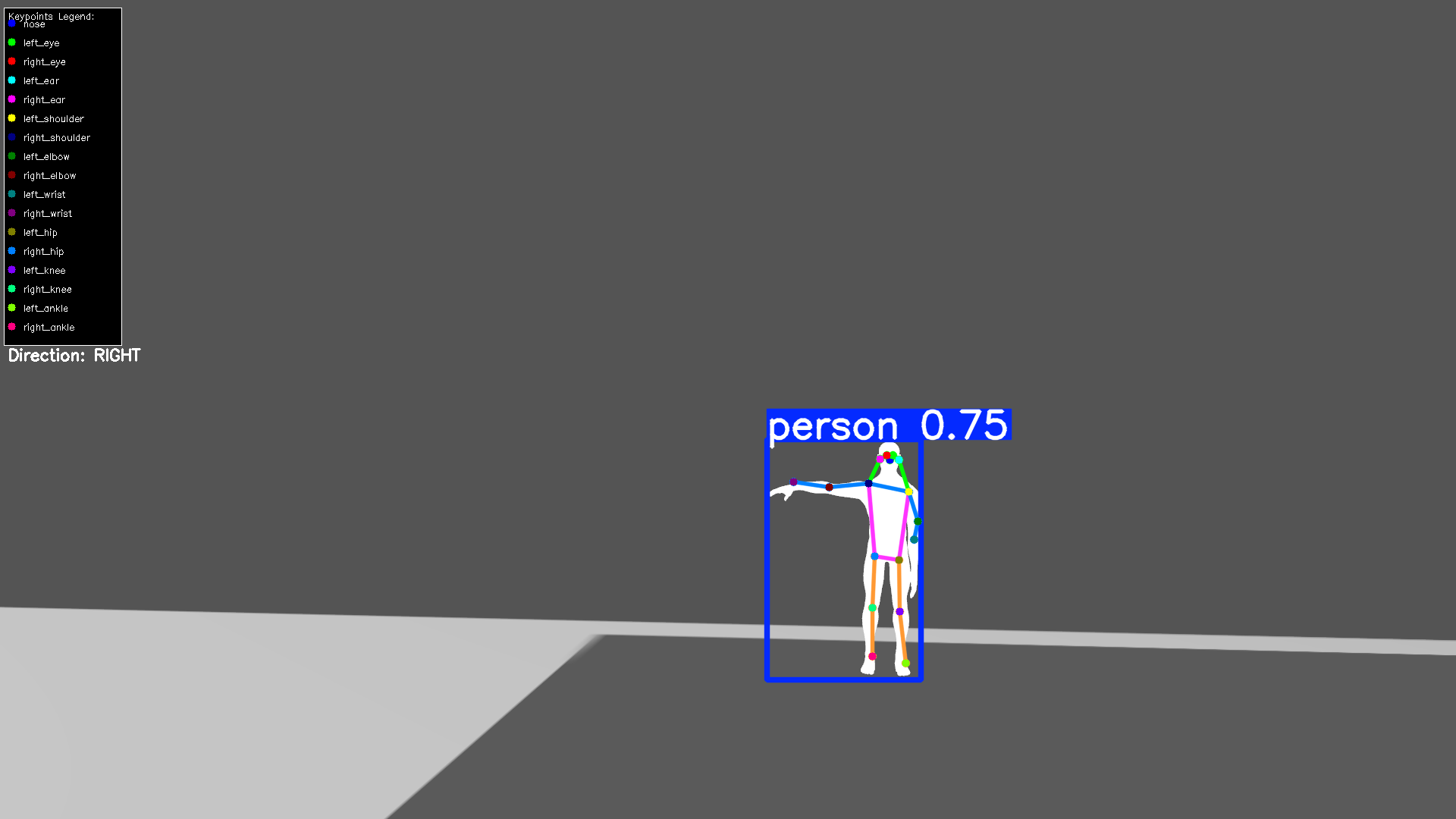}}\\
    \subfloat[\textit{(d)}]{\includegraphics[width=0.3\textwidth,height=0.18\textwidth]{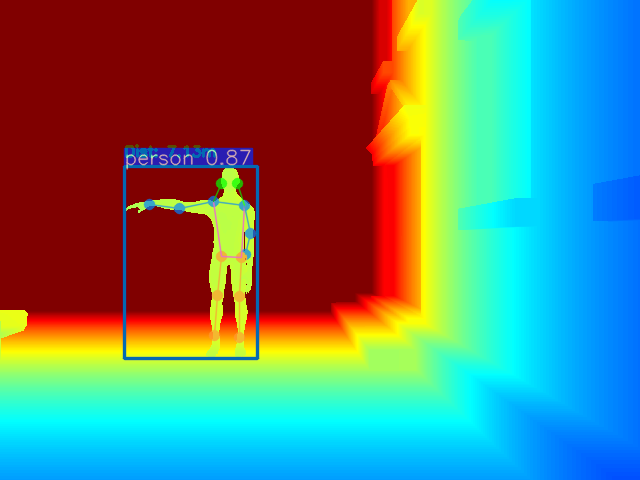}}\hfill
    \subfloat[\textit{(e)}]{\includegraphics[width=0.3\textwidth,height=0.18\textwidth]{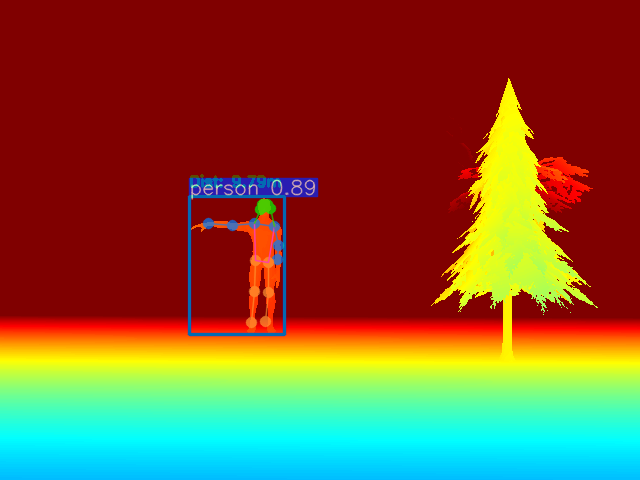}}\hfill
    \subfloat[\textit{(f)}]{\includegraphics[width=0.3\textwidth,height=0.18\textwidth]{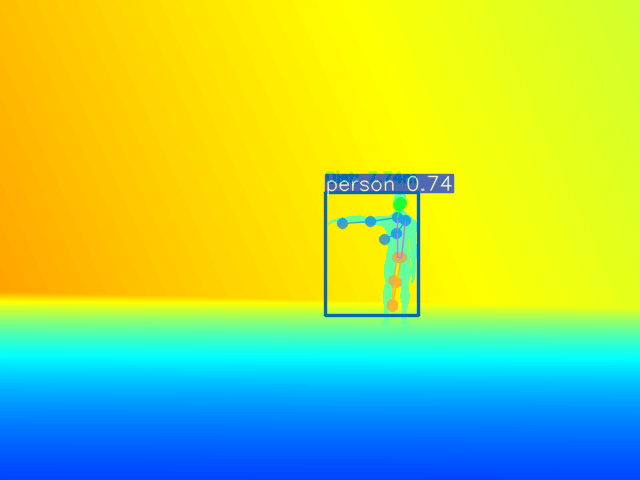}}\\
    \caption{(\textit{a,b,c}) are the RGB versions and (\textit{d,e,f}) are the Depth versions for City, Forest, and Wall scenarios, respectively, depicting the Closest Correct Estimation (CCE).}
    \label{fig:closest_correct_estimation}
\end{figure*}

CCE Estimation is a reliability metric for gesture recognition that measures how often a model correctly identifies the intended gesture in each frame. Its recognition reliability is assessed by calculating the proportion of frames where YOLOv11 correctly identifies the gesture, based on the nearest detection instance (Fig.~\ref {fig:closest_correct_estimation}).

\begin{equation}
\text{CCE} = \frac{1}{F} \sum_{f=1}^{F} \mathbf{1} \left( \hat{y}_f = y_f \right)
\end{equation}

\begin{itemize}
    \item \(F\) is the total number of frames in the evaluation set.
    \item \(y_f\) is the ground-truth gesture label in frame \(f\).
    \item \(\hat{y}_f\) is the predicted gesture label of the detection instance closest to the ground-truth in frame \(f\).
    \item \(\mathbf{1}()\) is the indicator function, which returns 1 if the condition is true (\(\hat{y}_f = y_f\)) and 0 otherwise.
\end{itemize}

Table~\ref{tab:keypoint_det_cce} summarises the CCE confidence \(\mathit{p}\) and depth distance \(\mathit{d}\) in each scenario.

\subsubsection{\textbf{City Scenario}} Under RGB‐only inference, CCE (\SI{0.83}{}) is high despite background distractions and urban clutter. However, depth-fused inference yields a slightly higher CCE (\SI{0.87}{}), especially when the distance is \SI{7.13}{\metre} from the target, indicating improved performance where depth features outperformed RGB by offering clearer spatial separation.

\subsubsection{\textbf{Forest Scenario}} RGB‐only moderate CCE (\SI{0.74}{}) improves significantly with depth (\SI{0.89}{}) when distance from target is \SI{9.79}{\metre}, which shows that texture similarity between humans and trees (e.g., brown bark vs. skin) makes it difficult for RGB models to differentiate them. Additionally, occlusion from dense foliage can hide parts of the human body, which RGB models struggle to predict without depth cues. Whereas depth information gives extra cues that help differentiate objects despite similar textures or occlusion.

\subsubsection{\textbf{Wall Scenario}} In wall scenario, the environment is minimal with fewer distractions, leading to similar performance for both RGB‐only (CCE = \SI{0.75}{}) and depth-fused (CCE = \SI{0.74}{}) at \SI{7.74}{\metre}. The simplicity of the environment reduces the need for depth information, as the wall provides consistent and clear features that both modalities can handle effectively.

So effectively, if we carefully notice in Table~\ref{tab:env_performance}, we can interpret that using depth fusion with RGB can increase the performance significantly.

\begin{table}[!t]
\centering
\caption{Impact of depth information (RGB-D) on Performance across Environments}
\label{tab:env_performance}
\begin{tabular}{lcccl}
\toprule
\textbf{Environment} & \textbf{Entropy} & \textbf{RGB-Only} & \textbf{RGB-D} \\
\midrule
Wall & 2.63 & 0.75 CCE & 0.74 (-1\%) \\
City & 5.66 & 0.83 CCE & \cellcolor{lightgreen}0.87 (+4\%) \\
Forest & 3.99 & 0.74 CCE & \cellcolor{lightgreen}0.89 (+15\%) \\
\bottomrule
\end{tabular}
\end{table}

\begin{figure*}[tp]
    \centering
    \subfloat[\textit{(a)}]{\includegraphics[width=0.3\textwidth,height=0.18\textwidth]{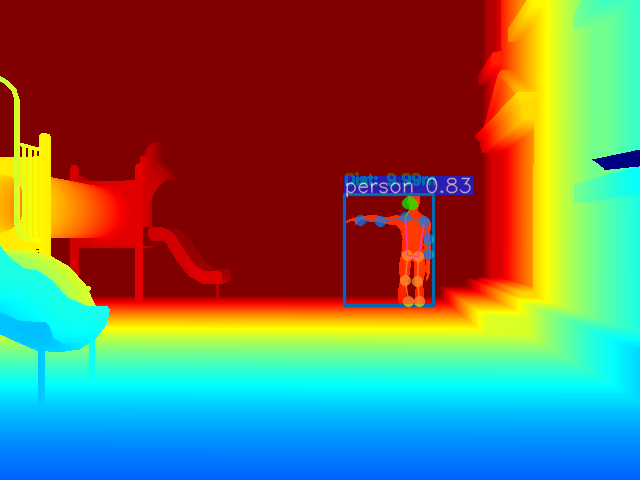}}\hfill
    \subfloat[\textit{(b)}]{\includegraphics[width=0.3\textwidth,height=0.18\textwidth]{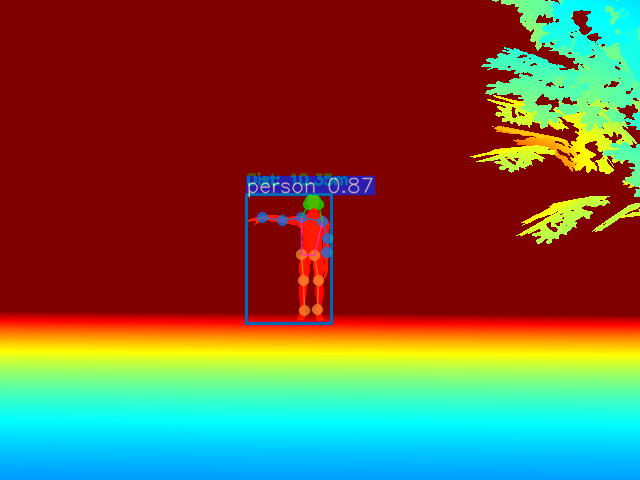}}\hfill
    \subfloat[\textit{(c)}]{\includegraphics[width=0.3\textwidth,height=0.18\textwidth]{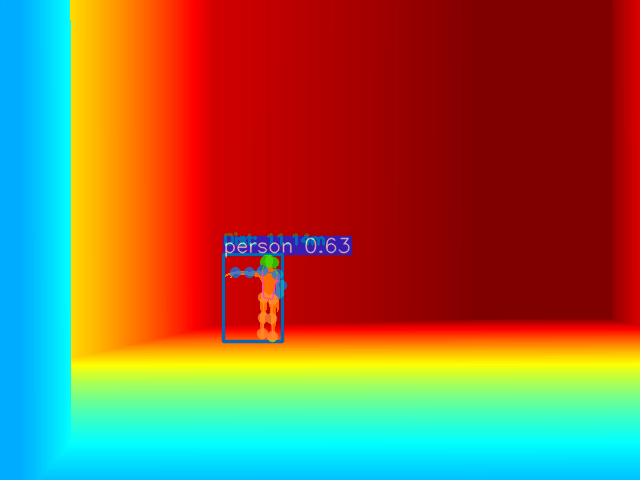}}\\
    \caption{(\textit{a,b,c}) show the RGB-based depth estimation results for City, Forest, and Wall scenarios, respectively.}
    \label{fig:depth}
\end{figure*}

\subsection{Depth Estimation Performance}
In this, we evaluated the performance of our YOLOv11-based detection system, enhanced with post-processing for 3D hand-gesture localisation, in estimating the 3D position of detected hand-gesture regions. Table~\ref{tab:yolo_depth_estimation} presents, for each scenario, the detection confidence \(\mathit{p}\) and the depth distance \(\mathit{d}\) from the reference distances of \SI{1}{\metre}, \SI{3}{\metre}, \SI{5}{\metre}, \SI{7}{\metre} and \SI{10}{\metre} from the camera (Fig.~\ref {fig:depth}).
We look for the absolute depth error, which is defined as the  magnitude of the difference between the reference and estimated depth:

\[
|\Delta d| = |d_{\text{ref}} - d_{\text{est}}| \quad \textit{(in metres)}
\]

\subsubsection{\textbf{City Scenario}} In the urban scene with mixed structures and foliage (Fig.~\ref{fig:depth}a), YOLOv11 maintains high confidence (\(\mathit{p}\) $\geq$ \SI{0.86}{}) up to \SI{3}{\metre}. Beyond \SI{3}{\metre}, confidence slightly degrades (\SI{0.77}{} at \SI{5}{\metre}, \SI{0.76}{} at \SI{7}{\metre}). The \(\Delta d\) ranges from \SI{0.01}{\metre} to \SI{0.24}{\metre} which is acceptable considering the behaviour of UAV.

\subsubsection{\textbf{Forest Scenario}} Under dense vegetation and dynamic occlusion (Fig.~\ref{fig:depth}b), confidence is lower at \SI{1}{\metre} (\SI{0.72}{}) with \(\Delta d\) = \SI{0.95}{\metre} indicating that being too close to the target in such a complex environment can cause significant estimation errors. However, at \SI{3}{\metre}, confidence rebounds to \SI{0.85}{}, likely due to improved foliage segmentation. At \SI{5}{\metre}, \SI{7}{\metre} and \SI{10}{\metre} confidence remains consistently high, which is \SI{0.86}{}, \SI{0.85}{} and \SI{0.87}{} respectively, indicating reliable detection through tree leaves and branches. The \(\Delta d\) ranges from \SI{0.20}{\metre} to \SI{0.35}{\metre}, which is reasonable given the complexity of the environment in which the UAV operates.

\subsubsection{\textbf{Wall Scenario}} In the enclosed corridor featuring a planar black wall (Fig.~\ref{fig:depth}c), the confidence at range from \SI{1}{\metre} to \SI{7}{\metre} generally gives good performance (\(\mathit{p}\) $\geq$ 0.78), thanks to uniform backgrounds that reduce visual clutter. However, confidence sporadically drops below (\(\mathit{p}\) $\leq$ 0.78) due to the lack of depth texture and minimal geometric variation, which leads to misalignment and less reliable depth estimates. Due to this, the absolute depth error (\SI{0.17}{\metre} $\leq$ \(\Delta d\) $\leq$ \SI{1.14}{\metre}) also remains higher.

\begin{table*}[tp]
    \centering
    \caption{Depth Estimation Results at Various Distances}
    \label{tab:yolo_depth_estimation}
    \begin{tabular}{ccccccccccc}
        \toprule
        & \multicolumn{2}{c}{\textbf{1 m}} & 
          \multicolumn{2}{c}{\textbf{3 m}} & 
          \multicolumn{2}{c}{\textbf{5 m}} & 
          \multicolumn{2}{c}{\textbf{7 m}} & 
          \multicolumn{2}{c}{\textbf{10 m}} \\
        \cmidrule(lr){2-3}
        \cmidrule(lr){4-5}
        \cmidrule(lr){6-7}
        \cmidrule(lr){8-9}
        \cmidrule(lr){10-11}
        \textbf{Scenario} & \(\mathit{p}\) & \(\mathit{d}\) & \(\mathit{p}\) & \(\mathit{d}\) & \(\mathit{p}\) & \(\mathit{d}\) & \(\mathit{p}\) & \(\mathit{d}\) & \(\mathit{p}\) & \(\mathit{d}\) \\
        \midrule
        \textbf{City}   & 0.86 & 1.24 & 0.87 & 3.11 & 0.77 & 4.81 & 0.76 & 7.01 & 0.83 & 9.99 \\
        \textbf{Forest} & 0.72 & 1.95 & 0.85 & 3.21 & 0.86 & 5.20 & 0.85 & 7.23 & 0.87 & 10.35 \\
        \textbf{Wall}   & 0.89 & 0.83 & 0.78 & 3.37 & 0.89 & 5.43 & 0.76 & 7.59 & 0.63 & 11.14 \\
        \bottomrule
    \end{tabular}
\end{table*}

\begin{figure*}[tp]
    \centering
    \subfloat[\textit{(a)}]{\includegraphics[width=0.3\textwidth]{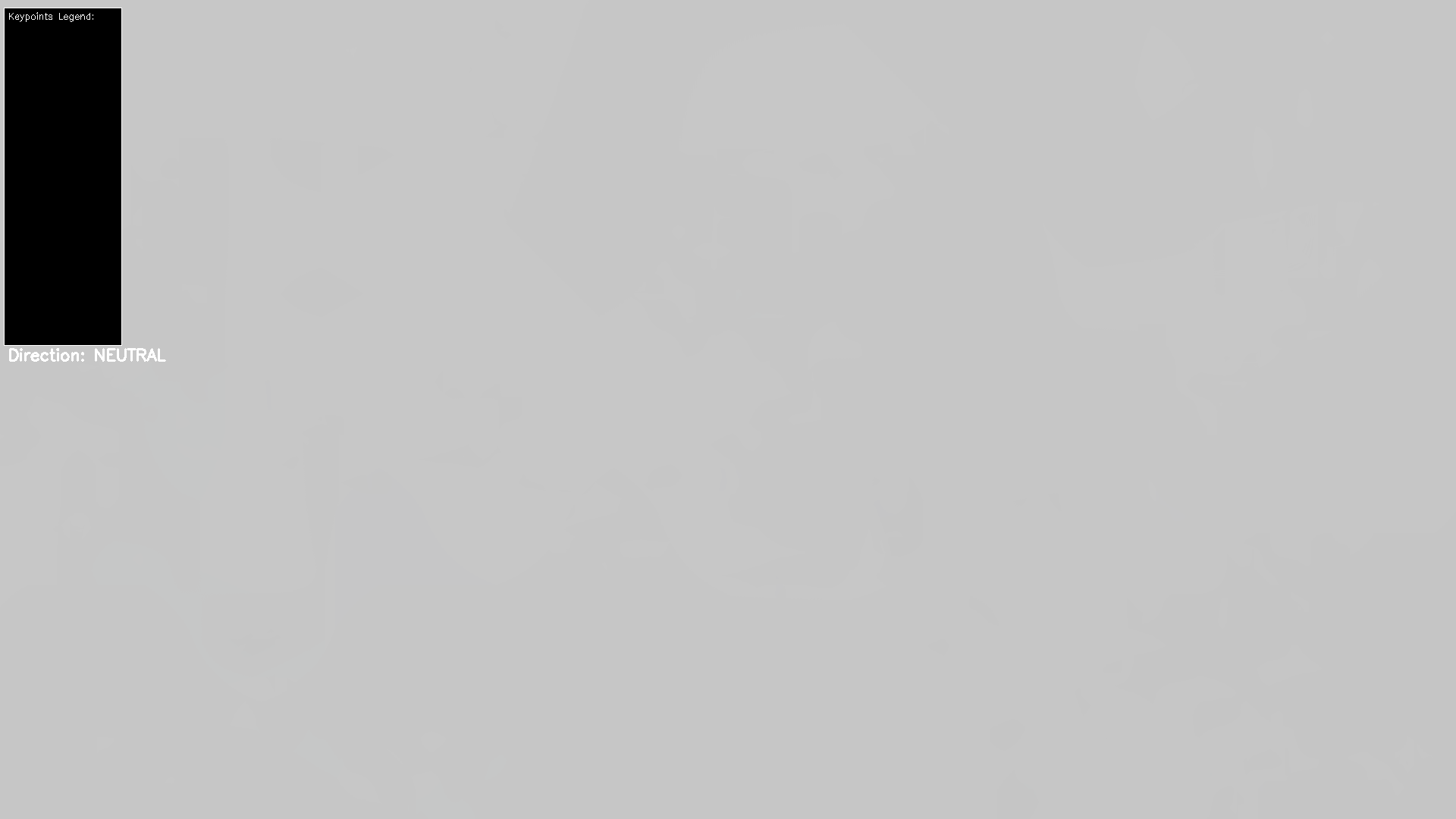}}\hfill
    \subfloat[\textit{(b)}]{\includegraphics[width=0.3\textwidth]{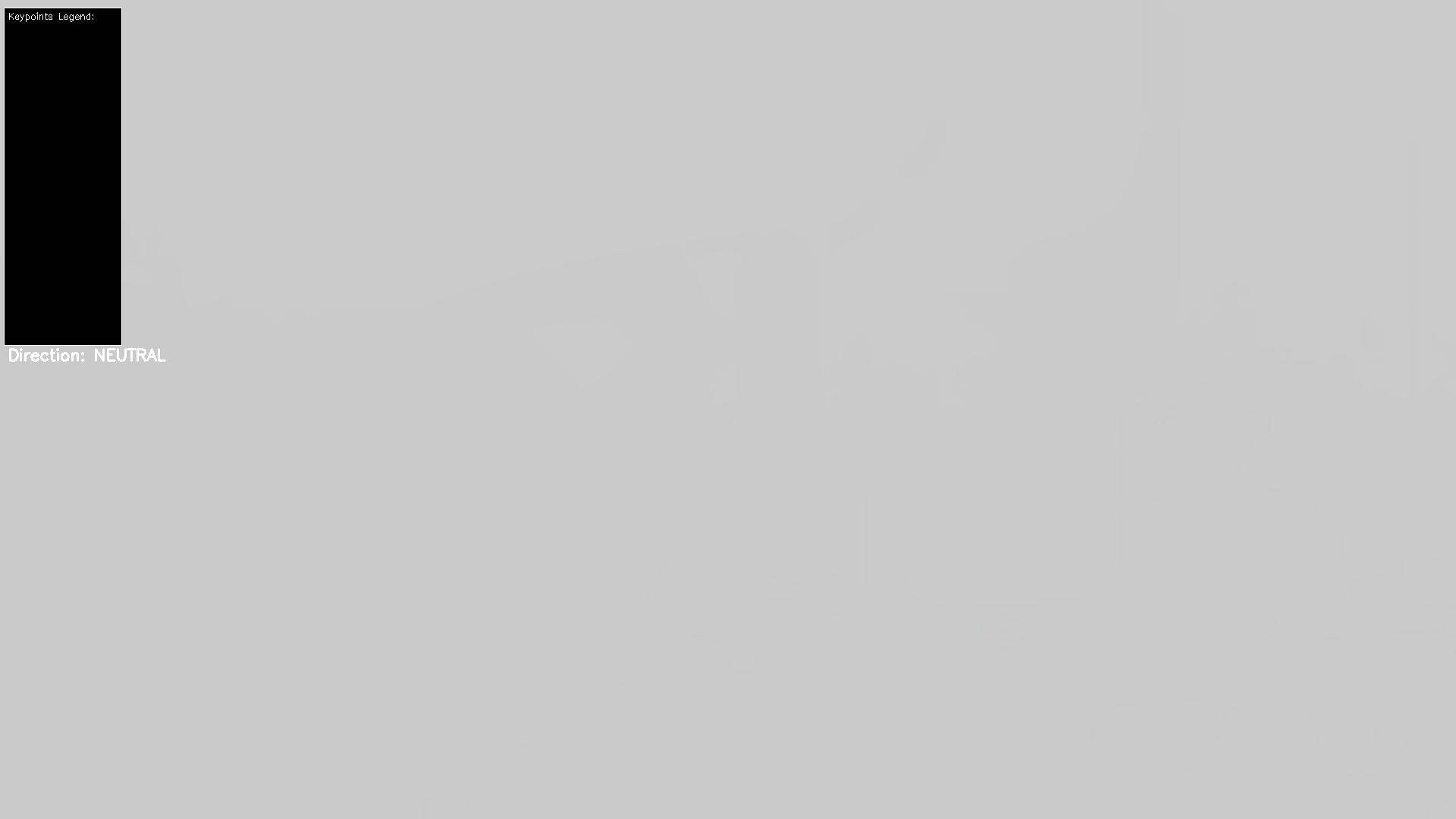}}\hfill
    \subfloat[\textit{(c)}]{\includegraphics[width=0.3\textwidth]{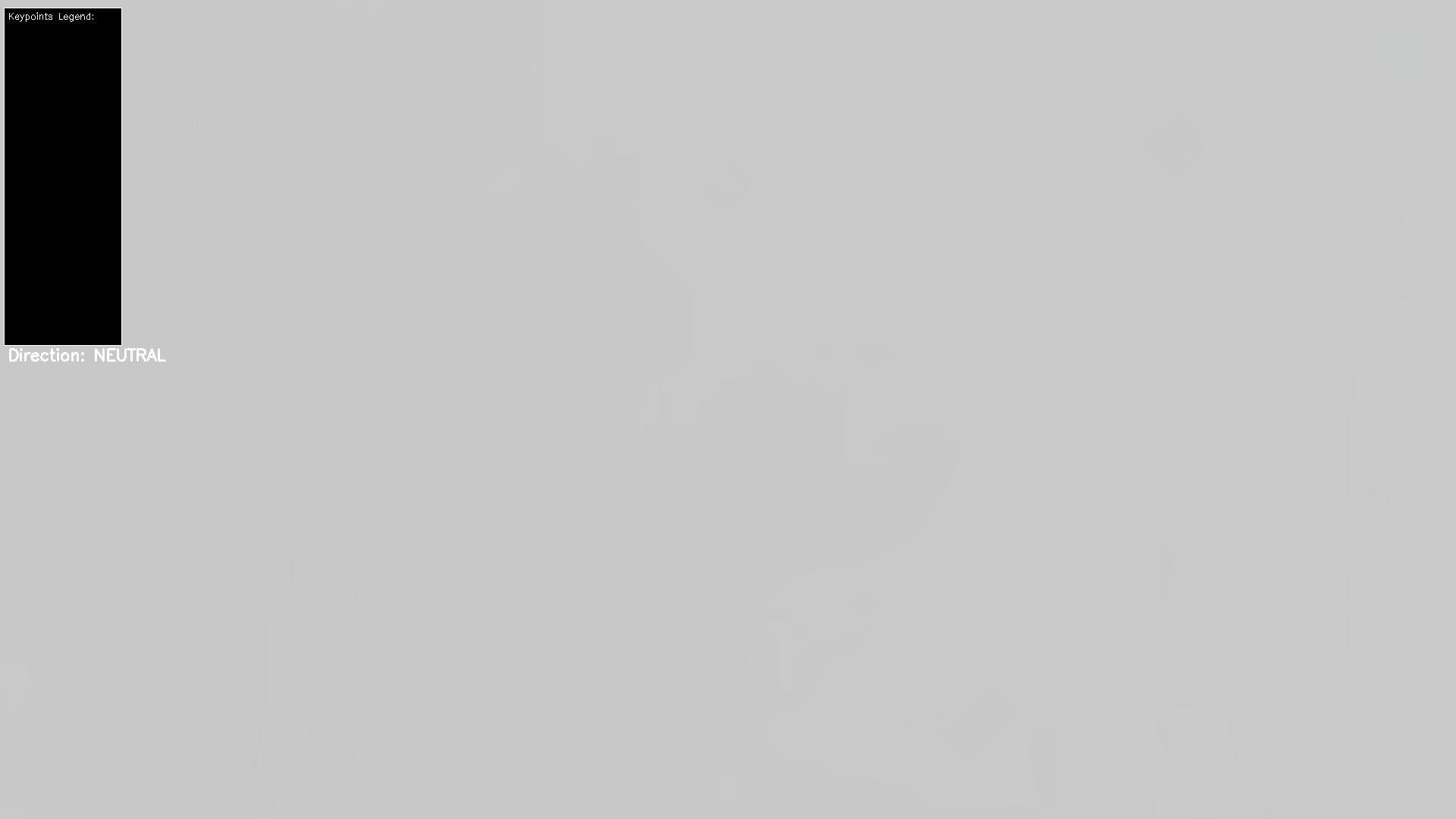}}\\[1ex]
    \subfloat[\textit{(d)}]{\includegraphics[width=0.3\textwidth]{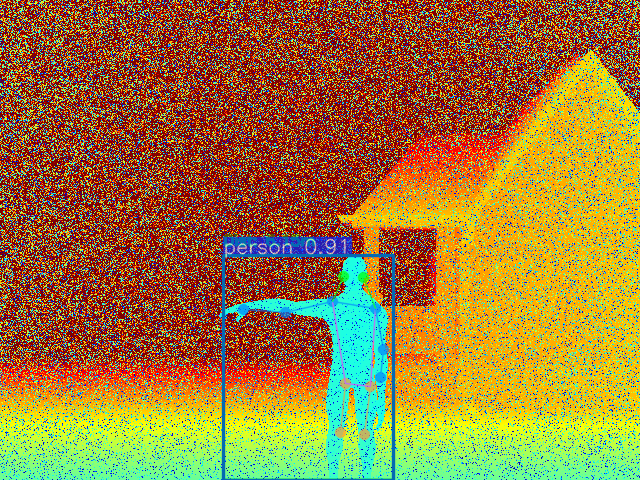}}\hfill
    \subfloat[\textit{(e)}]{\includegraphics[width=0.3\textwidth]{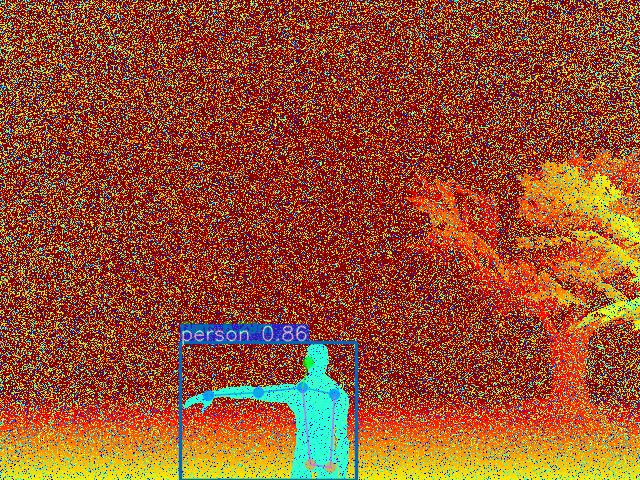}}\hfill
    \subfloat[\textit{(f)}]{\includegraphics[width=0.3\textwidth]{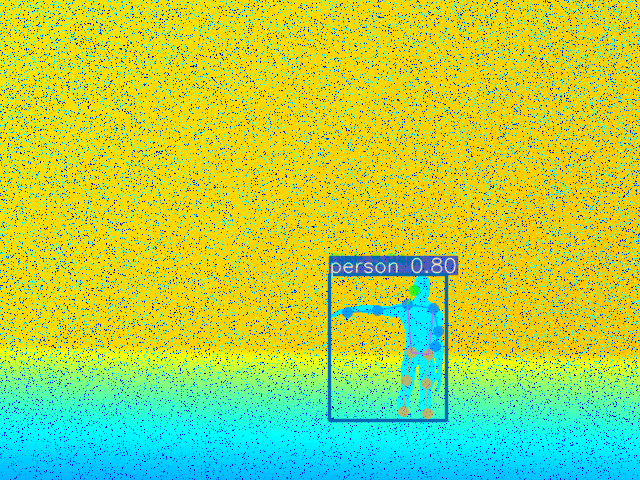}}\\
    \caption{(\textit{a, b, c}) are the RGB versions and (\textit{d, e, f}) are the depth versions for City, Forest and Wall scenarios, respectively, depicting the depth estimation under foggy conditions (farthest case).}
    \label{fig:depth_fog}
\end{figure*}

\begin{table*}[tp]
    \caption{Depth Estimation Results in Foggy Conditions}
    \label{tab:yolo_depth_fog}
    \centering
    \begin{tabular}{ccccccc}
        \toprule
        & \multicolumn{2}{c}{\textbf{Closest Distance}} & 
          \multicolumn{2}{c}{\textbf{Medium Distance}} & 
          \multicolumn{2}{c}{\textbf{Farthest Distance}} \\
        \cmidrule(lr){2-3}
        \cmidrule(lr){4-5}
        \cmidrule(lr){6-7}
        \textbf{Scenario} & \(\mathit{rgb}\) & \(\mathit{depth}\) & \(\mathit{rgb}\) & \(\mathit{depth}\) & \(\mathit{rgb}\) & \(\mathit{depth}\) \\
        \midrule
        \textbf{City}    & 0.28 & 0.59 & 0 & 0.88 & 0 & 0.91 \\
        \textbf{Forest}  & 0.63 & 0.60 & 0.30 & 0.30 & 0 & 0.86 \\
        \textbf{Wall}    & 0.57 & 0.45 & 0 & 0.60 & 0 & 0.80 \\
        \bottomrule
    \end{tabular}
\end{table*}

\begin{table*}[tp]
\centering
\caption{Fog Robustness Quantification of RGB and Depth-Based Detection}
\begin{tabular}{lccc}
\toprule
\textbf{Distance Category} & \textbf{RGB Confidence} & \textbf{Depth Confidence} & \textbf{Improvement} \\
\midrule
Close Range (1--3 m) & 0.28--0.63 & 0.45--0.60 & Moderate gain (mean +$\sim$25\%) \\
Medium Range (5 m) & 0.00--0.30 & 0.30--0.88 & Large gain (RGB often near zero) \\
\rowcolor{lightyellow}
Far Range (7--12 m) & 0.00 (fails) & 0.80--0.91 & Detection enabled (RGB fails) \\
\midrule
\multicolumn{4}{c}{\textbf{Overall Trend:} Depth sensing maintains reliable confidence as RGB degrades} \\
\bottomrule
\end{tabular}
\label{tab:fog_robustness}
\end{table*}

In contrast, the depth sensor remains substantially more resilient -- confidence exceeds \SI{0.80}{} at the farthest distance in both City (\SI{0.91}{}) and Wall (\SI{0.80}{}), and holds at \SI{0.86}{} in Forest despite foliage and atmospheric scattering. These findings underscore the importance of sensor fusion in adverse weather --- when RGB becomes unreliable, depth cues preserve detection capability --- our approach offers practical resilience in degraded visual environments, which is critical for real-world SAR operations under fog or smoke.
The overall quantification of fog robustness is presented in Table~\ref{tab:fog_robustness}

\subsection{Depth Estimation Performance in Foggy Conditions}

To evaluate robustness under visibility degradation and varied light intensity, we tested the YOLOv11 model in simulated foggy environments. Table~\ref{tab:yolo_depth_fog} presents detection confidence from both RGB and depth modalities across three scene types and distance bands -- closest, medium, and farthest (Fig.~\ref {fig:depth_fog}). As expected, the RGB-based detection performance deteriorates sharply in fog, with values dropping to zero at medium and far ranges (except for the medium-forest case) in all scenarios.\\ Even at the closest range, RGB confidence remains low (\SI{0.28}{} in City, \SI{0.63}{} in Forest, and \SI{0.57}{} in Wall), confirming that visual occlusion from fog severely impairs conventional camera input.

\subsection{Comparative Analysis and Insights}

Combining the above results yields the following key insights:

\subsubsection{\textit{Depth Fusion Enhances Robustness in Cluttered Scenes}} In the closest correct estimation task from section~\ref{sec:result_sec}, depth fusion along with RGB improves confidence by 4\% to 15\% in City and Forest settings, showing that depth input helps in cluttered and occluded scenes.
\subsubsection{\textit{Operational Range Limitations}} For SAR‐relevant human gesture-based guidance, practical detection ranges lie up to \SI{12}{\metre}, based on the specifications of the onboard camera module. Within this range, performance is further constrained in heavily occluded environments (e.g., forests), whereas detection remains more reliable in open or planar settings (e.g., urban areas or walls). Beyond this operational range, YOLOv11 confidence scores and depth estimation accuracy degrade significantly, indicating that mission planning should maintain operators within close proximity to the UAV for consistent gesture recognition.

\subsubsection{\textit{Environment‐Dependent CCE Behaviour}} The CCE reveals that RGB dominance can outperform depth in texture‐sparse scenarios (wall), while depth outperforms RGB in visually complex scenes (City). Thus, an adaptive, environment‐aware fusion policy—tuned in real time by measuring scene‐entropy—could further optimise gesture recognition.
The system monitors the scene in real time and assesses its visual complexity. If the scene is simple (like a plain wall), it prioritises using colour (RGB) information. While in more complex scenes (like a city street), it relies more heavily on depth information. This adaptive approach enhances the overall system’s ability.

\subsubsection{\textit{Validation Against Recent Literature}} A critical gap in recent SAR-specific literature is the lack of quantified performance in degraded visual environments. While Li et al. \cite{li_efficient_2024} optimised for model efficiency and Abdalla and Baidya \cite{abdalla_uav_2025} achieved 96.14\% accuracy in clear weather, neither study evaluated system failure rates in fog or smoke—conditions ubiquitous in post-disaster scenarios. Similarly, the YOLOv9 SAR ensemble \cite{incoft25} focused on detection accuracy (94.18\%) and did not quantify its performance in fog.

In contrast, our results quantify a 46.7\% average performance improvement when using depth fusion in foggy conditions compared to RGB-only baselines. In severe fog (visibility <5m), our RGB-only confidence dropped to near-zero, while the proposed depth-aware fusion maintained operational confidence levels (0.80–0.91). This establishes that while competitor systems may offer marginal gains in clear-weather peak accuracy, our framework provides essential operational reliability in low-visibility environments where purely RGB-based systems fail.
The results are presented in Table~\ref{tab:env_performance_comparision} and Table~\ref{tab:unified_comparison}

\subsubsection{\textit{Architectural Robustness: Skeleton-Based Estimation vs. Direct Detection}} A critical architectural distinction in this work is the choice of Direct Gesture Detection using a YOLO-based framework over skeleton-based pose estimation methods such as MediaPipe, AirPose, and VTTransPose. As shown in Table~\ref{tab:failure_mode_analysis}, while skeleton-based approaches can achieve high precision in clear conditions by explicitly modelling joint coordinates (e.g., elbows, wrists, and shoulders), they rely on sufficiently visible human figures and exhibit fundamental failure modes in SAR-specific degraded environments. In contrast, the detection-oriented architecture of YOLOv11 facilitates more robust human identification under long-range observation, partial occlusion, and challenging UAV operating conditions.
\begin{itemize}
    \item \textbf{The ``Graph Collapse'' Phenomenon in Fog:} Skeleton-based systems rely on the construction of a connected graph $G = (V, E)$, where $V$ represents visible keypoints. In heavy fog or smoke (visibility < 5m), high-frequency edge details required to resolve specific joints—such as a wrist or elbow—are often obliterated. If a critical node (e.g., the elbow) is obscured by smoke, the edge connections cannot be formed, causing the entire pose graph to collapse. This results in a ``null'' output, where the system fails to register a human presence despite the silhouette being visible.
    \item \textbf{Superiority of Volumetric ``Blob'' Detection:} In contrast, our Direct Detection framework models the operator’s gesture at the person level, using keypoints as supportive cues rather than as a rigid dependency chain. By leveraging fused RGB-D features, the system captures low-frequency structural descriptors -- such as the extended torso-arm silhouette -- which remain preserved even when high-frequency internal details (e.g., joint contours) are degraded by fog.
\end{itemize}

\begin{table*}[tp]
\centering
\caption{Competitor systems -- Generalised performance overview}
\begin{tabular}{lccccc}
\toprule
\textbf{System} & \textbf{Test Condition} & \textbf{Performance} & \textbf{Env-Specific?} & \textbf{Fog Testing} \\
\midrule
MediaPipe & Mobile, desktop & Not reported & Not detailed & NO \\
YOLOv9 SAR & Heridal dataset & 94.18\% mAP & Heridal-specific & NO \\
VTTransPose & COCO val2017 & 76.5\% AP & Generic COCO & NO \\
YOLOv11pose & Animal posture & 94.1\% mAP & Domain-specific & NO \\
Gesture YOLOv4 & Clear weather & 96.14\% acc & Optimal only & NO \\
\rowcolor{lightgreen}
Our Proposed System & Gazebo  & 89\% average & City, Forest and Wall & \textbf{YES} \\
\bottomrule
\end{tabular}
\label{tab:env_performance_comparision}
\end{table*}

\begin{table*}[tp]
\centering
\renewcommand{\arraystretch}{1.4}
\caption{Unified competitive comparison across detection, pose, and gesture-based systems. (Green denotes unique or best-in-class capabilities of the proposed system for UAV-based search and rescue).}
\label{tab:unified_comparison}

\resizebox{\textwidth}{!}{%
\begin{tabular}{|l|c|c|c|c|c|c|}
\hline
\rowcolor{lightblue}
\textbf{Dimension}
& \textbf{Proposed System} 
& \textbf{MediaPipe} 
& \textbf{YOLOv9 SAR} 
& \textbf{VTTransPose} 
& \textbf{YOLOv11} 
& \textbf{Gesture YOLOv4} \\
\hline

Task Focus 
& Gesture$\rightarrow$Command 
& Perception pipeline 
& Detection 
& Pose estimation 
& Detection 
& Gesture detection \\
\hline

Fog / Smoke Robustness 
& \cellcolor{lightgreen}+46.7\% 
& NO 
& NO 
& NO 
& NO 
& NO \\
\hline

Occlusion Handling 
& \cellcolor{lightgreen}+15\% 
& --
& Ensemble 
& --
& --
& -- \\
\hline

Adaptive RGB-D Fusion 
& \cellcolor{lightgreen}YES
& NO 
& NO 
& NO 
& NO 
& NO \\
\hline

Temporal Consistency 
& \cellcolor{lightgreen}YES 
& NO 
& NO 
& NO 
& NO 
& NO \\
\hline

SAR-Specific Design 
& \cellcolor{lightgreen}YES 
& NO 
& YES 
& NO 
& NO 
& NO \\
\hline

Multi-Environment Testing 
& \cellcolor{lightgreen}YES 
& NO 
& YES 
& NO 
& NO 
& NO \\
\hline

Hardware 
& RealSense D435i 
& Generic 
& Generic 
& Generic 
& Generic 
& Not specified \\
\hline

Platform 
& Gazebo + PX4 
& Mobile, desktop 
& Not specified 
& GPU lab 
& Edge GPU 
& Edge server \\
\hline

Real-Time Middleware 
& \cellcolor{lightgreen}ROS2 DDS
& Not specified 
& Not specified 
& Not specified 
& Not specified 
& Not specified \\
\hline

Latency 
& $<$50 ms 
& Not reported
& Not reported
& Not reported
& Not reported
& $<$30 ms \\
\hline

Operational Range (m) 
& \cellcolor{lightgreen}12 
& Not specified
& Not specified
& Not specified
& Not specified
& 10 \\
\hline

Dataset Type 
& Gazebo + Blender
& Not applicable 
& Heridal 
& COCO 
& COCO 
& Unknown \\
\hline
\end{tabular}
}
\end{table*}

\begin{table*}[tp]
\centering
\footnotesize
\caption{Failure Mode Analysis of Recognition Architectures in SAR Environments}
\label{tab:failure_mode_analysis}
\renewcommand{\arraystretch}{1.1}
\begin{tabular}{{p{3cm} p{4.5cm} p{4.5cm}}}
    \toprule
    \textbf{Feature} & \textbf{Skeleton Tracking} \newline (AirPose / MediaPipe) & \textbf{Direct Detection} \newline (Proposed System) \\
    \midrule
    \textbf{Detection Unit} & Discrete Keypoints (Joints) & Holistic Object \\
        
    \textbf{Dependency} & Requires high-resolution visibility of individual limbs & Requires general depth/shape profile \\
        
    \textbf{Fog Behaviour} & \textbf{Catastrophic Failure:} Graph breaks if a single joint is obscured & \textbf{Robust:} Interpolates intent from overall shape cues (+46.7\% gain) \\
        
    \textbf{Occlusion} & \textbf{High Jitter:} Lost tracking leads to noise and null outputs & \textbf{Stable:} Classifies visible volumetric features \\
        
    \textbf{SAR Viability} & \textbf{Limited:} Valid only in clear weather & \textbf{High:} Validated in fog and three different environments \\
        \bottomrule
    \end{tabular}
\end{table*}

\subsubsection{Evaluation Metrics and Statistical Analysis for Foggy Condition}

This work evaluates the effect of sensor modality on perception performance by comparing \textit{RGB-only} and \textit{RGB+Depth} configurations. Performance is quantified using \textbf{model confidence scores}, where higher values indicate stronger prediction certainty and improved perceptual reliability. Since confidence values are continuous and paired across identical environments and distance conditions, paired statistical metrics are employed to assess both statistical significance and effect magnitude.

\paragraph*{Paired Confidence Difference}:
For each experimental condition $i$, a paired confidence difference is computed as
\[
d_i = C^{\text{RGB+Depth}}_i - C^{\text{RGB}}_i
\]
where $C^{\text{RGB+Depth}}_i$ and $C^{\text{RGB}}_i$ denote confidence values obtained using RGB+Depth and RGB-only inputs, respectively. Positive values of $d_i$ indicate improved confidence due to depth fusion. All subsequent analyses are performed on the paired difference set $\{d_i\}_{i=1}^{N}$.

\paragraph*{Wilcoxon Signed-Rank Test}:
The Wilcoxon signed-rank test is a nonparametric statistical test used to assess whether the median of paired differences differs significantly from zero \cite{wilcoxon_individual_1945}. Unlike the paired $t$-test, it does not assume normality of the differences and is therefore suitable for small sample sizes and potentially skewed distributions.

Given that higher confidence values indicate better performance, the hypotheses are defined as:

\begin{align*}
H_0 &: \mathrm{median}(d_i) \le 0, \\
H_1 &: \mathrm{median}(d_i) > 0.
\end{align*}

The test ranks the absolute nonzero differences $|d_i|$, assigns the original sign to each rank, and computes the sum of signed ranks. Statistical significance is evaluated using a one-sided test with a significance level of $\alpha = 0.05$.

\paragraph*{Bootstrap Confidence Interval}:
While hypothesis testing determines whether a statistically significant improvement exists, it does not quantify the magnitude of the improvement. To estimate effect size and uncertainty, we compute bootstrap confidence intervals over the paired differences \cite{efron_bootstrap_1979, efron_introduction_1994}.

Given the set $\{d_i\}$, bootstrap resampling is performed by sampling $N$ values with replacement to form a bootstrap sample $\{d^{(b)}_i\}$. For each bootstrap sample $b$, a statistic of interest (mean or median) is computed as:

\[
\hat{\theta}^{(b)} = \mathrm{statistic}\left(d^{(b)}_1, \ldots, d^{(b)}_N\right).
\]

This process is repeated $B$ times (with $B \ge 10{,}000$), yielding an empirical bootstrap distribution.

\paragraph*{Percentile-Based Confidence Interval}:
Generally, we adopt a 95\% confidence level, which is standard in statistical analysis, as it provides a reliable balance between confidence and interval width and enables consistent comparison with prior work. A 95\% bootstrap confidence interval is obtained using the percentile method:

\[
\mathrm{CI}_{95\%} = \left[\hat{\theta}_{2.5\%}, \hat{\theta}_{97.5\%}\right].
\]

The values 2.5\% and 97.5\% arise from distributing the remaining probability mass equally between the two tails:

\[
\frac{100\% - 95\%}{2} = 2.5\%.
\]

Thus, the interval retains the central 95\% of the bootstrap distribution, providing an estimate of uncertainty without assuming any parametric form.

\paragraph*{Interpretation of Results}:
Applying the above metrics to confidence values under fog conditions from Table~\ref{tab:yolo_depth_fog} yields the following results:
\begin{itemize}
\item Mean confidence improvement: $0.47$
\item 95\% bootstrap confidence interval (mean): $[0.20,\ 0.73]$
\item Median confidence improvement: $0.60$
\item 95\% bootstrap confidence interval (median): $[-0.03,\ 0.88]$
\item Wilcoxon signed-rank test: $p = 0.0195$
\end{itemize}

The Wilcoxon test indicates a statistically significant improvement in confidence when using RGB+Depth ($p < 0.05$). The bootstrap confidence interval for the mean difference lies entirely above zero, demonstrating a consistent and meaningful improvement. Although the median confidence interval marginally overlaps zero, the positive median and predominance of positive paired differences indicate that depth fusion generally enhances confidence under fog conditions, with limited variability across environments.

By combining the Wilcoxon signed-rank test with bootstrap confidence intervals, we obtain a robust, distribution-free assessment of the improvement in confidence. The Wilcoxon test determines whether the improvement is statistically significant, while the bootstrap intervals estimate its magnitude and associated uncertainty. Together, they provide complementary evidence of the effectiveness of RGB+Depth fusion, particularly under challenging environmental conditions.

{
\section{Extended Evaluation}\label{sec:ext_eval}
This section extends the evaluation of the framework. The results of Section~\ref{sec:result_sec} were obtained with the stock YOLOv11n-pose (COCO weights \cite{lin_microsoft_2014}, without retraining) and Algorithm~\ref{alg:algo}. Here, we implement the modified YOLOv11 of Section~\ref{sec:meth_sec}, trained with the complete loss $\mathcal{L}_{total}$ of Eq.~(\ref{eq:loss_fun}) (YOLOv11-$\mathcal{L}_{total}$), and the depth-weighted pose estimation of Eq.~(\ref{eq:multimodal})--(\ref{eq:weights}), and we test them together with Algorithm~\ref{alg:algo}, on a simulated test set and on a dataset of real people, in clear air and in two fog models.

\subsection{Implementation of the Modified YOLOv11}\label{sec:t4_impl}
YOLOv11-$\mathcal{L}_{total}$ keeps every YOLOv11n-pose layer with its COCO weights, changes the classes of the anchor-free head to $LEFT$, $RIGHT$ and $NEUTRAL$, and adds four modules (Fig.~\ref{fig:yolo11arch}, Table~\ref{tab:t4_layers}):
\begin{itemize}
\item \textit{Depth-aware FPN with RGB-D fusion:} the input has four channels, B, G, R and the aligned depth $D$ (0.1--20~m mapped to 0--255). A small depth network $\phi$ computes depth features at the scales of the backbone outputs $P_3$, $P_4$ and $P_5$ (strides 8, 16 and 32, as in FPN \cite{lin_feature_2017}) and adds them after layers 4, 6 and 10:
\begin{equation}
\label{eq:fusion}
\tilde{P}_l = P_l + W_l\big(\phi_l(D)\big), \qquad l = 3, 4, 5,
\end{equation}
where $W_l$ is a $1\times1$ convolution whose weights start at zero, so that training starts from the RGB model. The fused features $\tilde{P}_l$ feed both the rest of the backbone and the neck.
\item \textit{Cross-scale attention:} the three neck outputs $N_3$, $N_4$ and $N_5$ (PANet \cite{liu_path_2018}) are re-weighted before the head:
\begin{equation}
\label{eq:attention}
\tilde{N}_l = N_l + \gamma_l \, A_l \odot N_l,
\end{equation}
where $A_l$ is an attention map with values between 0 and 1, computed as in CBAM \cite{woo_cbam_2018} from a channel weight (an MLP on the pooled features of all three scales) and a spatial weight (a $7\times7$ convolution over their mean and max maps), $\odot$ is the element-wise product, and $\gamma_l$ is a learned gain that starts at zero. As $A_l$ is computed for each frame, the fused RGB-D features are weighted adaptively.
\item \textit{Temporal consistency:} $\mathcal{L}_{temporal}$ (Eq.~(\ref{eq:l_temporal})) keeps the gesture scores of a person stable over consecutive frames during training.
\item \textit{Distance head:} predicts the distance of the person at every anchor point of the head, for $\mathcal{L}_{depth}$.
\end{itemize}

\textit{Custom Multi-Task Loss Function:} The loss of Eq.~(\ref{eq:loss_fun}) is implemented with the weights
\begin{equation}
\label{eq:loss_weights}
\begin{aligned}
\mathcal{L}_{total} ={}& \underbrace{7.5\,\mathcal{L}_{CIoU} + 1.5\,\mathcal{L}_{DFL}}_{\mathcal{L}_{box}} + \underbrace{0.5\,\mathcal{L}_{BCE}}_{\mathcal{L}_{conf} + \mathcal{L}_{gesture}} \\
& + \underbrace{12\,\mathcal{L}_{kpt} + 1.0\,\mathcal{L}_{kobj}}_{\text{keypoints}} + 1.0\,\mathcal{L}_{depth} + 1.0\,\mathcal{L}_{temporal}.
\end{aligned}
\end{equation}
$\mathcal{L}_{CIoU}$ \cite{zheng_distance-iou_2020} and $\mathcal{L}_{DFL}$ \cite{li_generalized_2020} measure the box error, $\mathcal{L}_{BCE}$ is the binary cross-entropy of the class scores, which covers $\mathcal{L}_{conf}$ and $\mathcal{L}_{gesture}$, and $\mathcal{L}_{kpt}$ and $\mathcal{L}_{kobj}$ \cite{maji_yolo-pose_2022} measure the keypoint position and visibility errors. The keypoint terms come from YOLOv11-pose and belong to the spatial loss of Eq.~(\ref{eq:loss_fun}). These five terms keep the default weights of YOLOv11 (Ultralytics 8.4.13). The two new terms have weight 1, as in Eq.~(\ref{eq:loss_fun}):
\begin{equation}
\label{eq:l_depth}
\mathcal{L}_{depth} = \frac{1}{n} \sum_{i=1}^{n} \left| \hat{d}_i - d_i \right|,
\end{equation}
\begin{equation}
\label{eq:l_temporal}
\mathcal{L}_{temporal} = \frac{1}{12} \sum_{f=1}^{4} \left\| \mathbf{g}_f - \bar{\mathbf{g}} \right\|^2 .
\end{equation}
$\mathcal{L}_{depth}$ is the mean absolute error between the predicted distance $\hat{d}_i$ and the true distance $d_i$ (m) over the $n$ anchor points that YOLOv11 matches to a person. $\mathcal{L}_{temporal}$ uses clips of four consecutive frames with the same gesture: $\mathbf{g}_f$ are the three gesture scores (logits) of the person in frame $f$, $\bar{\mathbf{g}}$ is their mean over the clip, and dividing by $4\times3 = 12$ gives the mean squared deviation. It is averaged over the two clips of each training step.

\textit{Computational cost:} YOLOv11-$\mathcal{L}_{total}$ has 3.52~M parameters (+22.4\% over YOLOv11n-pose) and needs 10.0 instead of 7.5 GFLOPs per $640\times640$ frame. Its median processing time in the offline tests (Section~\ref{sec:t4_eval}) was 50~ms per frame, against 19~ms for the stock YOLOv11n-pose with Algorithm~\ref{alg:algo}.

\textit{Depth-weighted pose estimation:} Eq.~(\ref{eq:multimodal})--(\ref{eq:weights}) are implemented for the stock YOLOv11n-pose without retraining. YOLOv11n-pose runs on the RGB frame and on the colour-mapped depth frame, and for each joint $j$ of the operator, $\vec{RGB}_j$ and $\vec{D}_j$ are its two keypoints placed in 3D with the measured depth. $\vec{T}_j$ is the mean of the joint's RGB and depth estimate over the last five frames, and $\vec{C}_j$ places the joint at the operator's torso distance, as the arms stay near the torso plane. The variance of $\vec{RGB}_j$ and $\vec{D}_j$ grows with low keypoint confidence and uneven depth, that of $\vec{T}_j$ with motion, and that of $\vec{C}_j$ is fixed at 0.25~m. $\lambda = 3000$ was chosen on the synthetic validation worlds (Section~\ref{sec:training}). The fused joints are projected back to the image, where Algorithm~\ref{alg:algo} decides the direction. As YOLOv11n-pose runs twice, the median processing time was 72~ms per frame.
}

\begin{table*}[tp]
\centering

\caption{Modules and parameters of YOLOv11-$\mathcal{L}_{total}$ for a $640\times640$ input (layer-by-layer in Fig.~\ref{fig:yolo11arch}). Conv = convolution, batch normalisation and SiLU; s2 = stride 2; sizes are channels $\times$ height$^2$.}
\label{tab:t4_layers}
\footnotesize
\begin{adjustbox}{max width=\textwidth}
\begin{tabular}{p{3.1cm} p{6.6cm} p{4.9cm} r}
\toprule
\textbf{Module} & \textbf{Layers} & \textbf{Output} & \textbf{Parameters} \\
\midrule
\multicolumn{4}{l}{\textit{YOLOv11n-pose, unchanged (COCO weights)}} \\
Backbone (L0--L8) & Conv $3\times3$ s2, C3k2 & $P_3$: $128\times80^2$ (L4); $P_4$: $128\times40^2$ (L6) & 951,136 \\
SPPF, C2PSA (L9, L10) & spatial pyramid pooling; partial self-attention & $P_5$: $256\times20^2$ & 414,336 \\
Neck (L11--L22) & upsampling, concatenation, C3k2, Conv s2 (FPN + PAN) & $N_3$: $64\times80^2$; $N_4$: $128\times40^2$; $N_5$: $256\times20^2$ & 793,696 \\
Pose head (L23) & per scale: box ($4\times16$ DFL bins; 381,904), 3 gesture classes (49,353), 17 keypoints (284,427) & 8,400 anchors & 715,684 \\
\midrule
\multicolumn{4}{l}{\textit{Added modules}} \\
Depth stem & Conv $3\times3$ s2 $1\rightarrow16$; Conv $3\times3$ s2 $16\rightarrow32$; Conv $3\times3$ $32\rightarrow32$ & $32\times160^2$ & 14,128 \\
Depth level $\phi_3$ & Conv $3\times3$ s2 $32\rightarrow64$; Conv $3\times3$ $64\rightarrow64$ & $64\times80^2$ & 55,552 \\
Depth level $\phi_4$ & Conv $3\times3$ s2 $64\rightarrow96$; Conv $3\times3$ $96\rightarrow96$ & $96\times40^2$ & 138,624 \\
Depth level $\phi_5$ & Conv $3\times3$ s2 $96\rightarrow128$; Conv $3\times3$ $128\rightarrow128$ & $128\times20^2$ & 258,560 \\
Fusion $W_3$, $W_4$, $W_5$ & $1\times1$ conv $64\rightarrow128$, $96\rightarrow128$, $128\rightarrow256$ (zero-initialised), added after L4, L6, L10 & as $P_3$, $P_4$, $P_5$ & 53,760 \\
Cross-scale attention & channel MLP $448\rightarrow56\rightarrow448$; per scale $7\times7$ conv $4\rightarrow1$; gains $\gamma_l$ (zero-initialised) & as $N_3$, $N_4$, $N_5$ & 51,274 \\
Distance head & per scale: Conv $3\times3$ $C_l\rightarrow16$; Conv $3\times3$ $16\rightarrow16$; $1\times1$ conv $16\rightarrow1$ & 8,400 distances & 71,667 \\
\midrule
\multicolumn{3}{l}{Total, YOLOv11n-pose with 3 classes (7.5 GFLOPs)} & 2,874,852 \\
\multicolumn{3}{l}{Total, YOLOv11-$\mathcal{L}_{total}$ (10.0 GFLOPs)} & 3,518,417 \\
\bottomrule
\end{tabular}
\end{adjustbox}
\end{table*}

{
\subsection{Dataset and Training}\label{sec:training}
YOLOv11-$\mathcal{L}_{total}$ was retrained twice with the same recipe, once on synthetic data only (synthetic-trained) and once on real people only (real-trained). Table~\ref{tab:t4_data} summarises both datasets.
\begin{itemize}
\item \textit{Synthetic dataset:} rendered in procedurally generated Gazebo worlds with randomly posed humans at 0.8--12~m. The City, Forest and Wall worlds were not used for training and are kept for testing only. Depth comes from the simulated OakD-Lite or RealSense D435i (with its noise model). In addition, 35\% of the clear frames get Koschmieder fog (Section~\ref{sec:fog_models}) with a visibility of 4--60~m.
\item \textit{Real dataset:} UAV-GESTURE \cite{perera_uav-gesture_2019}, split by subject ID, with depth estimated from RGB by Depth-Anything-V2 \cite{yang_depth_2024}. As the dataset has no distance ground truth, $\mathcal{L}_{depth}$ is inactive for the real-trained model.
\item \textit{Gesture definitions and annotation:} $LEFT$ or $RIGHT$ means that arm is raised sideways (shoulder 60--110$^\circ$, elbow bent at most 40$^\circ$) while the other arm is down (below 35$^\circ$). $NEUTRAL$ means both arms are down or both are raised. Labels and keypoints are computed from the posed skeleton (synthetic) or the dataset's joint annotations (real), without manual labelling.
\item \textit{Training:} from the COCO weights (added modules from scratch) for 100 epochs, batch 16, $640\times640$, FP32, seed 0, keeping the best validation epoch. The optimiser was selected automatically: MuSGD with a learning rate of 0.01 (synthetic) and AdamW with 0.00143 (real), both with momentum 0.9, weight decay 0.0005 and 3 warm-up epochs. Augmentation used HSV jitter of the colour channels (0.015, 0.7, 0.4, depth unchanged), translation by $\pm10\%$, scaling by $\pm50\%$ and mosaic (off in the last 10 epochs). Random horizontal flipping was disabled, because it mirrors the image without swapping the $LEFT$ and $RIGHT$ labels, and rotation, shear and mixup were not used. Instead, every real training frame was also stored mirrored with $LEFT$ and $RIGHT$ swapped, to balance the two classes. Training took 5.4~h (synthetic) and 2.9~h (real) on an NVIDIA RTX 4000 SFF Ada (20+20~GB) with an Intel Xeon Silver 4310, Ultralytics 8.4.13 and PyTorch 2.4.1.
\end{itemize}
}

\begin{table*}[tp]
\centering

\caption{Training data of YOLOv11-$\mathcal{L}_{total}$. Class counts are labelled persons ($LEFT$ / $RIGHT$ / $NEUTRAL$).}
\label{tab:t4_data}
\footnotesize
\begin{adjustbox}{max width=\textwidth}
\begin{tabular}{p{3.2cm} p{6.4cm} p{6.4cm}}
\toprule
& \textbf{Synthetic (synthetic-trained)} & \textbf{Real, UAV-GESTURE (real-trained)} \\
\midrule
Source & 24 procedurally generated Gazebo worlds (12 clear, 12 particle fog) & 119 videos of 13 marshalling signals, filmed from a hovering UAV \\
Subjects & 150 generated humans: 120 training, 30 validation & 8 participants, released as 11 subject IDs: training S4, S6, S9, S12, S15 (53 videos); validation S5, S13 (26); test S1, S3, S11, S14 (40) \\
Training images & 8,000 (20 worlds $\times$ 400) & 4,840 (every 6th frame, each also mirrored with $LEFT$/$RIGHT$ swapped) \\
Validation images & 1,200 (4 worlds $\times$ 300) & 1,818 \\
Classes, training & 2,136 / 2,160 / 2,207; 1,497 images without a person & 302 / 302 / 4,236 \\
Classes, validation & 332 / 308 / 321; 239 images without a person & 60 / 60 / 1,698 \\
Clips for $\mathcal{L}_{temporal}$ & 735 training clips of 4 frames & 464 training clips of 4 frames (3 video frames apart) \\
Depth & Simulated OakD-Lite or RealSense D435i (with noise model) & Estimated from RGB (Depth-Anything-V2) \\
Test data & City, Forest, Wall and their fog versions (Section~\ref{sec:t4_eval}) & all 11 subject IDs (synthetic-trained); S1, S3, S11, S14 (real-trained) (Section~\ref{sec:real_eval}) \\
\bottomrule
\end{tabular}
\end{adjustbox}
\end{table*}

{
\subsection{Fog Models}\label{sec:fog_models}
We have used two fog models. One is the \textit{Particle fog} which is the Gazebo particle emitter fog of Sections~\ref{sec:sim_sec}--\ref{sec:result_sec}, where the RGB camera renders it, and it scatters the returns of the depth camera. As it exists only inside the simulator, it cannot be added to recorded videos. For the real videos, and for a similar comparison on the simulated dataset, we therefore also use \textit{Koschmieder fog} (Section~\ref{sec:rw_ext}), as Sakaridis et al. \cite{sakaridis_semantic_2018} did for real images, where each pixel of an RGB frame is attenuated with the transmission $t = e^{-3.912\,r/V}$ and blended with an airlight of 217 grey levels, where $r$ is the distance of the pixel (ground truth in simulation, Depth-Anything-V2 estimate on real frames) and $V$ = 50, 20, 10 or 5~m is the visibility. The depth image loses all returns where $t < 2\%$, that is, beyond $V$. Koschmieder fog is therefore homogeneous and set by the visibility, whereas the particle fog is dense and patchy and hides the person from about 5--10~m onwards.
}

{
\subsection{Evaluation in Simulation}\label{sec:t4_eval}
The test set uses the City, Forest and Wall worlds, which were used neither for training nor for choosing $\lambda$. The three test human models show $LEFT$, $RIGHT$ and $NEUTRAL$ at 1, 3, 5, 7 and 10~m, each seen from eight UAV positions (360 frames per condition, 720 for particle fog), in clear air, particle fog and Koschmieder fog with $V$ = 50--5~m. A prediction is correct when the direction of the selected person equals the gesture, and a false command is a $LEFT$ or $RIGHT$ output for a $NEUTRAL$ gesture (Table~\ref{tab:t4_sim}).

Without any training, Algorithm~\ref{alg:algo} reaches 71\% on RGB and 61\% on depth in clear air. In fog, RGB drops to 28\% in particle fog and to 43\% and 11\% at $V$ = 10 and 5~m, whereas depth keeps 36\%, 55\% and 27\% respectively. As in Section~\ref{sec:result_sec}, RGB is the more accurate and more confident input in clear air (mean confidence 0.71 vs.\ 0.50), and depth is the more robust one in fog. At 7--10~m in particle fog, however, the person is also lost in the depth image, with a mean confidence of 0.00--0.14 against 0.80--0.91 at the farthest range in Table~\ref{tab:yolo_depth_fog}. The advantage of depth in fog is therefore smaller here than in Section~\ref{sec:result_sec}.

The fused pose of Eq.~(\ref{eq:multimodal})--(\ref{eq:weights}) combines both inputs and reaches 76\% in clear air and 41\% in particle fog. It is better than RGB in every condition (McNemar's test \cite{mcnemar_note_1947}, $p < 0.05$) and better than depth in every condition except $V$ = 10 and 5~m, where the differences are not significant, with at most 5\% false commands. YOLOv11-$\mathcal{L}_{total}$, trained on synthetic data, reaches 91\% in clear air and 86--88\% in particle fog and down to $V$ = 10~m, but it issues more false commands (11--21\%).

In six closed-loop flights in Gazebo with PX4 SITL and ROS2, covering the three worlds in clear air and fog, YOLOv11-$\mathcal{L}_{total}$ issued the command 0.33--0.66~s after each gesture change, and the UAV started turning after 1.03--1.30~s.
}

\begin{table*}[tp]
\centering

\caption{Direction accuracy (\%) on the simulated test set (360 frames per condition, 720 for particle fog), with false commands (\%) in brackets. $LEFT$/$RIGHT$ accuracy is for clear air.}
\label{tab:t4_sim}
\footnotesize
\setlength{\tabcolsep}{4.5pt}
\begin{adjustbox}{max width=\textwidth}
\begin{tabular}{l l l c c c c c c c}
\toprule
& & & & & \multicolumn{4}{c}{\textbf{Koschmieder fog, $V$}} & \\
\cmidrule(lr){6-9}
\textbf{Pipeline} & \textbf{Input} & \textbf{Trained on} & \textbf{Clear} & \textbf{Particle fog} & \textbf{50 m} & \textbf{20 m} & \textbf{10 m} & \textbf{5 m} & \textbf{\shortstack{$LEFT$/$RIGHT$\\accuracy}} \\
\midrule
Algorithm~\ref{alg:algo} & RGB & none & 71 (4) & 28 (0) & 76 (0) & 68 (0) & 43 (0) & 11 (0) & 60 \\
Algorithm~\ref{alg:algo} & Depth & none & 61 (0) & 36 (0) & 61 (0) & 61 (0) & 55 (0) & 27 (0) & 54 \\
Algorithm~\ref{alg:algo}, fused pose & RGB-D & none & 76 (5) & 41 (0) & 79 (0) & 77 (0) & 57 (0) & 26 (0) & 67 \\
YOLOv11-$\mathcal{L}_{total}$ & RGB-D & synthetic & 91 (11) & 86 (13) & 91 (11) & 91 (13) & 88 (21) & 50 (14) & 93 \\
\bottomrule
\end{tabular}
\end{adjustbox}
\end{table*}

{
\subsection{Evaluation on Real People}\label{sec:real_eval}
We used UAV-GESTURE \cite{perera_uav-gesture_2019}, which contains 119 videos of eight participants, recorded on two days and released as 11 subject IDs, filmed from a hovering UAV at about 6.5--8.4~m (estimated). Every 3rd frame was labelled from the dataset's joint annotations with the gesture definitions of Section~\ref{sec:training}, and its depth was estimated from RGB. As almost all pointing frames show $RIGHT$, every frame was also tested mirrored, with $LEFT$ and $RIGHT$ swapped. Table~\ref{tab:real} reports all 11 subject IDs, none of which was used for Algorithm~\ref{alg:algo}, the choice of $\lambda$ or the synthetic-trained model, and separately the four subject IDs held out from the real-trained model, whose participants do not appear in its training IDs (checked visually).

Without any training, Algorithm~\ref{alg:algo} reads the gestures of real people with 80\% accuracy on RGB and on estimated depth (79--80\% mirrored), and most of its errors are false commands (17--19\%). In Koschmieder fog, RGB holds up to $V$ = 20~m but drops to 24\% at $V$ = 10~m, whereas depth keeps 77\%. At $V$ = 5~m, which is shorter than the person's distance, both inputs lose the person. This confirms on real people that depth is the more robust input in fog, although the depth here was estimated from the clear frame.

The fused pose reaches 78\% in clear air, 1.9 points below RGB and depth (95\% confidence interval against RGB over subject IDs, bootstrap \cite{efron_introduction_1994}: $[-3.0, -0.8]$). At $V$ = 10~m it reaches 77\%, which is 53 points above RGB and on par with depth. On real people, the estimated depth carries no information independent of RGB, and the temporal term lags gesture changes by up to five frames. The fusion therefore does not improve on the better single input as it does in simulation.

YOLOv11-$\mathcal{L}_{total}$ trained on synthetic data reaches only 68\% (62\% mirrored), with 33\% false commands. It learned the gestures from rendered humans, whereas the stock keypoints were learned from real people (COCO) and the angle thresholds of Algorithm~\ref{alg:algo} do not depend on appearance. Trained on five other subject IDs, it reaches 99\% (100\% mirrored) with almost no false commands (0.1\%), but it drops to 89\% at $V$ = 20~m and fails at $V$ = 10~m, because its real training data contained no fog. The modified YOLOv11 is therefore effective when it is trained on data from the target domain, including its weather. Without such data, Algorithm~\ref{alg:algo} with the stock model is the more robust choice.
}

\begin{table*}[tp]
\centering

\caption{Direction accuracy (\%) on real people (UAV-GESTURE), with false commands (\%) in brackets. $LEFT$/$RIGHT$ and mirrored accuracy are for clear air. Depth is estimated from RGB.}
\label{tab:real}
\footnotesize
\setlength{\tabcolsep}{4.5pt}
\begin{adjustbox}{max width=\textwidth}
\begin{tabular}{l l l c c c c c c c}
\toprule
& & & & \multicolumn{4}{c}{\textbf{Koschmieder fog, $V$}} & \multicolumn{2}{c}{\textbf{Clear air}} \\
\cmidrule(lr){5-8}\cmidrule(lr){9-10}
\textbf{Pipeline} & \textbf{Input} & \textbf{Trained on} & \textbf{Clear} & \textbf{50 m} & \textbf{20 m} & \textbf{10 m} & \textbf{5 m} & \textbf{\shortstack{$LEFT$/$RIGHT$\\accuracy}} & \textbf{Mirrored} \\
\midrule
\multicolumn{10}{l}{\textit{All 11 subject IDs: 10,006 frames (8,727 $NEUTRAL$, 1,245 $RIGHT$, 34 $LEFT$)}} \\
Algorithm~\ref{alg:algo} & RGB & none & 80 (19) & 79 (20) & 77 (19) & 24 (1) & 0 (0) & 75 & 80 \\
Algorithm~\ref{alg:algo} & Depth (est.) & none & 80 (17) & 80 (17) & 80 (17) & 77 (18) & 0 (0) & 82 & 79 \\
Algorithm~\ref{alg:algo}, fused pose & RGB-D (est.) & none & 78 (20) & 78 (21) & 75 (22) & 77 (18) & 0 (0) & 68 & 78 \\
YOLOv11-$\mathcal{L}_{total}$ & RGB-D (est.) & synthetic & 68 (33) & 69 (33) & 69 (33) & 71 (29) & 13 (14) & 90 & 62 \\
\midrule
\multicolumn{10}{l}{\textit{Four held-out subject IDs (S1, S3, S11, S14): 3,361 frames (2,811 $NEUTRAL$, 550 $RIGHT$)}} \\
Algorithm~\ref{alg:algo} & RGB & none & 78 (20) & 78 (21) & 78 (16) & 8 (0) & 0 (0) & 69 & 79 \\
Algorithm~\ref{alg:algo} & Depth (est.) & none & 79 (18) & 79 (18) & 79 (18) & 76 (17) & 0 (0) & 78 & 78 \\
Algorithm~\ref{alg:algo}, fused pose & RGB-D (est.) & none & 77 (20) & 77 (21) & 77 (19) & 73 (18) & 0 (0) & 60 & 77 \\
YOLOv11-$\mathcal{L}_{total}$ & RGB-D (est.) & synthetic & 63 (36) & 64 (37) & 67 (36) & 68 (32) & 3 (3) & 79 & 57 \\
YOLOv11-$\mathcal{L}_{total}$ & RGB-D (est.) & real, 5 IDs & 99 (0) & 99 (0) & 89 (0) & 1 (0) & 0 (0) & 95 & 100 \\
\bottomrule
\end{tabular}
\end{adjustbox}
\end{table*}

{
\subsection{Limitations}
\begin{itemize}
\item The UAV was flown only in simulation (Gazebo, PX4 SITL), not on physical hardware.
\item Fog is simulated (particle fog, Koschmieder's law) and cannot reproduce all effects of real fog or smoke. On real people, it was applied with estimated depth.
\item On real people, depth was estimated from RGB rather than measured by a depth sensor, so the depth input carries no information independent of RGB. UAV-GESTURE also differs from SAR scenes in background, viewpoint and gestures.
\item In Eq.~(\ref{eq:multimodal}), the contextual prior $\vec{C}_j$ is realised as a body-plane prior, and $\vec{T}_j$ acts only on video, not on the single frames of the simulated test set.
\item YOLOv11-$\mathcal{L}_{total}$ needs training data from the target domain, including its weather conditions, to transfer to real people.
\end{itemize}
}

\section{Conclusion and Future Work}
\label{sec:conc_sec}

This work presents a novel HITL UAV framework that improves adaptability in unstructured environments by integrating gesture-based human guidance in SAR missions. The framework is developed on ROS2, focusing on long-term maintainability and scalability for SAR operations. The framework utilises a YOLOv11-based gesture recognition module to ensure robust detection in cluttered environments. Experimental results from PX4-Gazebo simulations show 89\% depth estimation confidence from the UAV's perspective, demonstrating its effectiveness. Depth-enhanced perception significantly reduces false positives and misclassifications in cluttered and dynamic SAR environments, while RGB remains valuable in texture‐poor but visually uniform settings. These results demonstrate that, under foggy conditions, depth information provides a significantly more reliable signal than RGB, with an average improvement of 46.7\% (from Table~\ref{tab:yolo_depth_fog}) across all evaluated scenarios. These findings provide quantitative guidance for deploying human-in-the-loop UAVs in real-world rescue missions, highlighting both the strengths and limitations of multimodal fusion under varied environmental conditions. The framework further demonstrates superior occlusion handling, achieving a 15\% improvement over RGB-only input. Adaptive RGB-D fusion combined with temporal consistency modelling enhances stable gesture interpretation and reduces perception noise in dynamic SAR scenarios.

The extended evaluation (Section~\ref{sec:ext_eval}) tested the framework beyond the three simulated scenes. Without any training, Algorithm~\ref{alg:algo} reads the pointing gestures of real people in the UAV-GESTURE dataset with 80\% accuracy, and in fog with a visibility of 10~m, depth keeps 77\% where RGB drops to 24\%, which confirms on real people that depth is the more robust input in fog. Fusing RGB and depth keypoints with Eq.~(\ref{eq:multimodal})--(\ref{eq:weights}) raises the accuracy in simulation (76\% in clear air and 41\% in particle fog, against 71\% and 28\% for RGB), but not on real people, where the depth is estimated from RGB. The modified YOLOv11, trained with $\mathcal{L}_{total}$, reaches 91\% in simulation and 99\% on unseen real people when trained on real data, but transfers poorly from simulation alone and fails in fog it was not trained on. Since the framework is modular, with separate ROS2 nodes for perception, gesture interpretation and control, its components can be replaced or retrained for other UAV tasks, such as other gesture vocabularies, cameras or missions, and evaluated on further datasets in the same way.

For Future scope, we can expand gesture vocabularies to improve detection models and support more nuanced, natural human interactions. The attention-driven RGB-D fusion of the modified YOLOv11 (Section~\ref{sec:ext_eval}) can be trained on data with extreme environmental degradation to further enhance robustness. Also, we can consider a field deployment to extend testing to real-world SAR missions to validate robustness in diverse environmental conditions beyond simulation.

In particular, real RGB-D gesture data recorded from UAVs in SAR-like conditions, including fog and smoke, are needed to train the modified YOLOv11 and to validate the depth-weighted pose estimation with a real depth sensor before field deployment.

\noindent{\color[gray]{0.4}\rule{\columnwidth}{1.5pt}}

\subsection*{Abbreviations}
{\footnotesize
\begin{tabular}{@{}p{2.1cm}p{6.0cm}@{}}
AdamW & Adam optimiser with decoupled weight decay \\
AP & average precision \\
BCE & binary cross-entropy \\
BGR8 & 8-bit blue, green and red image format \\
C2PSA & cross stage partial block with spatial attention \\
C3k2 & cross stage partial block with kernel size 2 \\
CBAM & convolutional block attention module \\
CCE & closest correct estimation \\
CI & confidence interval \\
CIoU & complete intersection over union \\
COCO & Common Objects in Context (dataset) \\
COLLADA & collaborative design activity (3D model format) \\
CPU & central processing unit \\
DDS & data distribution service \\
DFL & distribution focal loss \\
FP32 & 32-bit floating point \\
FPN & feature pyramid network \\
GFLOPs & giga floating-point operations \\
GPU & graphics processing unit \\
HITL & human-in-the-loop \\
HSV & hue, saturation and value \\
IMU & inertial measurement unit \\
IoU & intersection over union \\
LTS & long-term support \\
mAP & mean average precision \\
MAVROS & MAVLink (Micro Air Vehicle Link) interface for ROS \\
MLP & multilayer perceptron \\
MuSGD & optimiser combining Muon and stochastic gradient descent \\
NaN & not a number \\
OS & operating system \\
PAN, PANet & path aggregation network \\
PDD & pose estimation, distance calculation and direction classification (Algorithm~\ref{alg:algo}) \\
QoS & quality of service \\
RGB & red, green and blue (colour image) \\
RGB-D & RGB and depth \\
ROI & region of interest \\
ROS, ROS1, ROS2 & Robot Operating System, versions 1 and 2 \\
RTPS & real-time publish-subscribe \\
SAR & search and rescue \\
SDF & Simulation Description Format \\
SiLU & sigmoid linear unit \\
SITL & software-in-the-loop \\
SPPF & spatial pyramid pooling, fast \\
UAV & unmanned aerial vehicle \\
YOLO & You Only Look Once \\
\end{tabular}}

\subsection*{Data Availability Statement}
The real-people evaluation uses the public UAV-GESTURE dataset \cite{perera_uav-gesture_2019}. The source code of the framework (ROS2 nodes, Gazebo worlds and human models, the modified YOLOv11 and its training script, the depth-weighted pose estimation and the evaluation scripts) is available at \url{https://github.com/anubhavml/UAVFramework_SAR}.

\subsection*{Author Contributions Statement}
The first three authors conceived the idea and carried out the implementation. The remaining authors contributed by verifying the results and reviewing the manuscript. All authors reviewed and approved the final version of the manuscript.

\subsection*{Funding Statement}
No funding was received for this work.

\subsection*{Conflict of Interest Disclosure}
The authors declare that they have no conflict of interest.

\bibliographystyle{unsrtnat}
\bibliography{wileyNJD-Chicago}

\end{document}